\documentclass{article}
\usepackage{grffile}

\usepackage{times}
\usepackage{graphicx} 
\usepackage{subfigure} 
\usepackage{amsmath}
\usepackage{amssymb}
\usepackage{natbib}

\usepackage{algorithm}
\usepackage{algorithmic}
\usepackage{afterpage}

\allowdisplaybreaks
\usepackage{hyperref}

\newcommand\numberthis{\addtocounter{equation}{1}\tag{\theequation}}

\newcommand{\inner}[1]{\left\langle#1\right\rangle}

\def\R{\mathbb{R}}

\def\N{\mathbb{N}}

\newcommand{\norm}[1]{\left\|#1\right\|}

\def\tr{\mathop{\rm tr}\nolimits}

\newcommand{\Id}{\mathbb{I}}

\def\argmin{\mathop{\rm arg\,min}\limits}

\def\diag{\mathrm{diag}}

\def\eps{\epsilon}

\def\min{\mathop{\rm min}\nolimits}
\def\max{\mathop{\rm max}\nolimits}

\newtheorem{lemma}{Lemma}[section]

\newtheorem{theorem}{Theorem}[section]
\newtheorem{corollary}{Corollary}[section]
\newtheorem{definition}{Definition}[section]

\newenvironment{proof}{\par\noindent{\bf Proof:\ }}{\hfill$\Box$\\[2mm]}

\newif\ifpaper
\papertrue 
\usepackage[accepted]{icml2017} 

\icmltitlerunning{Variants of RMSProp and Adagrad with Logarithmic Regret Bounds}

\begin{document} 

\twocolumn[
\icmltitle{
Variants of RMSProp and Adagrad with Logarithmic Regret Bounds}




\begin{icmlauthorlist}
\icmlauthor{Mahesh Chandra Mukkamala}{uni1,uni2}
\icmlauthor{Matthias Hein}{uni1}
\end{icmlauthorlist}

\icmlaffiliation{uni1}{Department of Mathematics and Computer Science, Saarland University, Germany}
\icmlaffiliation{uni2}{IMPRS-CS, Max Planck Institute for Informatics, Saarbr\"ucken, Germany }

\icmlcorrespondingauthor{Mahesh Chandra Mukkamala}{mmahesh.chandra873@gmail.com}

\icmlkeywords{deep learning, online learning, machine learning, ICML, Online convex optimization, Non-convex optimization, Adagrad, RMSProp, SC-Adagrad, SC-RMSProp, Stochastic Optimization, Stochastic Gradient Descent, SGD}

\vskip 0.3in
]



\printAffiliationsAndNotice{}  

\begin{abstract} 
Adaptive gradient methods have become recently very popular, in particular as they have
been shown to be useful in the training of deep neural networks. In this paper we have analyzed
RMSProp, originally proposed for the training of deep neural networks, in the context of online convex
optimization and show $\sqrt{T}$-type regret bounds. Moreover, we propose two variants SC-Adagrad
and SC-RMSProp for which we show logarithmic regret bounds for strongly convex functions.
Finally, we demonstrate in the experiments that these new variants outperform other adaptive gradient techniques
or stochastic gradient descent in the optimization of strongly convex functions as well as in
training of deep neural networks.
\end{abstract} 


\section{Introduction}\label{sec:intro}
There has recently been a lot of work on adaptive gradient algorithms such as Adagrad \citep{duchi2011adaptive},  RMSProp \cite{hinton2012lecture}, ADADELTA \cite{Zei2012}, and Adam \citep{kingma2014adam}. The original idea of Adagrad to have a parameter specific learning rate by analyzing the gradients observed during the optimization turned out to be useful not only
in online convex optimization but also for training deep neural networks. The original analysis of Adagrad \citep{duchi2011adaptive}
was limited to the case of all convex functions for which it obtained a data-dependent regret bound of order $O(\sqrt{T})$ which
is known to be optimal \cite{Haz2016} for this class. However, a lot of learning problems have more structure in the sense that 
one optimizes over the restricted class of strongly convex functions. It has been shown in \citep{hazan2007logarithmic} that one
can achieve much better logarithmic regret bounds for the class of strongly convex functions.

The goal of this paper is twofold. First, we propose SC-Adagrad which is a variant of Adagrad adapted to the strongly convex case.
We show that SC-Adagrad achieves a logarithmic regret bound for the case of strongly convex functions, which is data-dependent.
It is known that such bounds can be much better in practice than data independent bounds  \citep{hazan2007logarithmic},\citep{mcmahan2014survey}. Second,
we analyze RMSProp which has become one of the standard methods to train neural networks beyond stochastic gradient descent.
We show that under some conditions on the weighting scheme of RMSProp, this algorithm achieves a data-dependent $O(\sqrt{T})$ regret bound. In fact, it turns out that RMSProp contains Adagrad as a special case  for a particular choice of the weighting scheme. Up to our knowledge this is the first theoretical result justifying the usage of RMSProp in online convex optimization and thus can at least be seen
as theoretical support for its usage in deep learning. Similarly, we then propose the variant SC-RMSProp for which we also show a
data-dependent logarithmic regret bound similar to SC-Adagrad for the class of strongly convex functions. Interestingly, SC-Adagrad
has been discussed in \cite{Rud2016}, where it is said that ``it does not to work''.  The reason for this is that SC-Adagrad comes along with a damping factor which prevents potentially large steps in the beginning of the iterations. However, as our analysis shows this damping factor has to be rather large initially to prevent large steps and should be then monotonically decreasing as a function of the iterations in order to stay adaptive. 
 Finally, we show in experiments on three datasets that the
new methods are competitive or outperform other adaptive gradient techniques as well as stochastic gradient descent for
strongly convex optimization problem in terms of regret and training objective but also perform very well in the training of deep neural networks, where we show results for different networks and datasets.

\section{Problem Statement}

We first need some technical statements and notation and then introduce the online convex optimization problem.
\subsection{Notation and Technical Statements}
We denote by $[T]$ the set $\{1,\ldots,T\}$.
Let $A \in \R^{d\times d}$ be a symmetric, positive definite matrix. We denote as
\begin{align*}
 \inner{x,y}_A &= \inner{x,Ay} = \sum_{i,j=1}^d A_{ij} x_i y_j, \quad \norm{x}_A &=\sqrt{\inner{x,x}_A} 
\end{align*}
Note that the standard Euclidean inner product becomes $\inner{x,y}=\sum_i x_i y_i = \inner{x,y}_\Id$
While we use here the general notation for matrices for comparison to the literature. All positive definite matrices $A$ in this paper
will be diagonal matrices, so that the computational effort for computing inner products and norms is still linear in $d$.
The Cauchy-Schwarz inequality becomes, $\inner{x,y}_A \, \leq \, \norm{x}_A \norm{y}_A.$
We further introduce the element-wise product $a \odot b$ of two vectors. Let $a,b \in \R^d$, then $(a \odot b)_i = a_i b_i$ for 
$i=1,\ldots,d$.

Let $A \in \R^{d \times d}$ be a symmetric, positive definite matrix, $z \in \R^d$ and $C \subset \R^d$ a convex set. 
Then we define the \textbf{weighted projection} $P_C^A(z)$ of $z$ onto the set $C$ as 
\begin{align}\label{eq:wproj}
P_C^A(z) = \argmin_{x \in C} \norm{x-z}^2_A.
\end{align}
It is well-known that the weighted projection is unique and non-expansive.
\begin{lemma}\label{le:nonexp}
Let $A \in \R^{d \times d}$ be a symmetric, positive definite matrix and $C\subset \R^d$ be a convex set. Then
\[ \norm{P^A_C(z)-P_C^A(y)}_A \, \leq \, \norm{z-y}_A.\]
\end{lemma}
\ifpaper
\begin{proof}
The first order optimality condition for the weighted projection in \eqref{eq:wproj} is given as
\[ A (x-z) \in N_C(x),\]
where $N_C(x)$ denotes the normal cone of $C$ at $x$. This can be rewritten as
\[ \inner{z-x,y-x}_A \, \leq \,0 \quad \forall y \in C.\]
This yields
\begin{align*}
& \inner{z-P_C^A(z),P_C^A(y)-P_C^A(z)}_A \leq 0,\\
& \inner{y-P_C^A(y),P_C^A(z)-P_C^A(y)}_A \leq 0.
\end{align*}
Adding these two inequalities yields
\begin{align*}
& \inner{z-P_C^A(z)-y+P_C^A(y),P_C^A(y)-P_C^A(z)}_A \leq 0 \\
& \Longrightarrow \; \norm{P_C^A(y)-P_C^A(z)}_A^2 \leq \inner{z-y,P_C^A(y)-P_C^A(z)}_A.
\end{align*}
The result follows from the application of the weighted Cauchy-Schwarz inequality. 
\end{proof}
\fi
\begin{lemma}
\label{lem:trace_prop}
For any symmetric, positive semi-definite matrix $A \in \mathbb{R}^{d\times  d}$ we have
\begin{equation}
\inner{x,Ax} \leq \lambda_{max}(A)\inner{x,x}\leq \tr(A)\inner{x,x}
\end{equation}
where $\lambda_{max}(A)$ is the maximum eigenvalue of matrix $A$ and $\tr(A)$ denotes the trace of matrix $A$\,.
\end{lemma}
\subsection{Problem Statement}
\label{sec:problem}
In this paper we analyze the online convex optimization setting, that is we have a convex set $C$ and at each round we get access to a (sub)-gradient of some continuous convex function $f_t: C \rightarrow \R$.
At the $t$-th iterate we predict $\theta_t \in C$ and suffer a loss $f_t(\theta_t)$. The goal is to perform well with respect to the optimal decision in hindsight defined as
\[ \theta^* = \argmin_{\theta \in C} \sum_{t=1}^T f_t(\theta).\]
The adversarial regret at time $T \in \N$ is then given as
\[ R(T)=\sum_{t=1}^T (f_t(\theta_t)-f_t(\theta^*)).\]
We assume that the adversarial can choose from the class of convex functions on $C$, for some parts we will specialize this to the set of strongly convex functions. 
\begin{definition}
Let $C$ be a convex set. We say that a function $f:C \rightarrow \R$ is $\mathbf{\mu}$-strongly convex, if
there exists $\mu \in \R^d$ with $\mu_i>0$ for $i=1,\ldots,d$ such that for all $x,y \in C$,
\begin{align*}
 f(y) \,&  \geq \,f(x) + \inner{\nabla f(x),y-x} + \norm{y-x}^2_{\diag(\mu)}\\
                &= f(x) + \inner{\nabla f(x),y-x} + \sum_{i=1}^d \mu_i (y_i-x_i)^2.
\end{align*}
Let $\zeta=\min_{i=1,\ldots,d} \mu_i$, then this function is $\zeta$-strongly convex (in the usual sense), that is
\[ f(y) \,\geq \, f(x) + \inner{\nabla f(x),y-x} + \zeta \norm{x-y}^2.\]
\end{definition}
Note that the difference between our notion of component-wise strong convexity and the usual definition of strong convexity
is indicated by the bold font versus normal font.
We have two assumptions:
\begin{itemize}
\item \textbf{A1:} It holds $\sup_{t\geq 1} \norm{g_t}_2 \leq G$ which implies the existence of a constant $G_\infty$ such that $\sup_{t\geq 1}\norm{g_t}_\infty \leq G_\infty$.
\item \textbf{A2:} It holds $\sup_{t \geq 1}\norm{\theta_t - \theta^*}_2 \leq D$ which implies the existence of a constant $D_\infty$ such that $\sup_{t\geq 1}\norm{\theta_t - \theta^*}_\infty \leq D_\infty$.
\end{itemize}

One of the first methods which achieves the optimal regret bound of $O(\sqrt{T})$ for convex problems is online projected gradient descent \cite{zinkevich2003online}, defined as
\begin{equation}
\theta_{t+1} = P_C(\theta_t - \alpha_t g_t)
\end{equation}
where $\alpha_t = \frac{\alpha}{\sqrt{t}}$ is the step-size scheme and $g_t$ is a (sub)-gradient of $f_t$ at $\theta_t$.  With $\alpha_t = \frac{\alpha}{t}$, online projected gradient descent method achieves the optimal $O(\log(T))$ regret bound for strongly-convex problems \citep{hazan2007logarithmic}. We consider Adagrad in the next subsection which is one of the popular adaptive alternative to online projected gradient descent.
\subsection{Adagrad for convex problems}
In this section we briefly recall the main result for the Adagrad. The algorithm for Adagrad is given in Algorithm \ref{alg:adagrad}.
\begin{algorithm}[tb]
   \caption{Adagrad}
   \label{alg:adagrad}
\begin{algorithmic}
   \STATE {\bfseries Input:} $\theta_1 \in C$\,,\,$\delta > 0 , v_0 = 0 \in \mathbb{R}^d$
   \FOR{$t=1$ {\bfseries to} T}
   \STATE $g_t \in \partial f_t(\theta_t)$
   \STATE $v_t = v_{t-1}+(g_t \odot g_t)$
   \STATE $A_t = \diag(\sqrt{v_t}) + \delta\Id$
   \STATE $\theta_{t+1} = P_C^{A_t}\big(\theta_t - \alpha A_t^{-1}g_t\big)$
   \ENDFOR
\end{algorithmic}
\end{algorithm}
If the adversarial is allowed to choose from the set of all possible convex functions on $C \subset \R^d$, then Adagrad achieves the
regret bound of order $O(\sqrt{T})$ as shown in  \citep{duchi2011adaptive}. This regret bound is known to be optimal for this class, see e.g.  \cite{Haz2016}. For better comparison to our results for RMSProp, we recall the result from \citep{duchi2011adaptive} in our notation.
For this purpose, we introduce the notation, $g_{1:T,i} = (g_{1,i},g_{2,i},..,g_{T,i})^T$, where $g_{t,i}$ is the $i$-th component of the 
gradient $g_t \in \R^d$ of the function $f_t$ evaluated at $\theta_t$.
\begin{theorem}\cite{duchi2011adaptive}\label{th:adagrad}
Let Assumptions \textbf{A1}, \textbf{A2} hold and let ${\theta_t}$ be the sequence generated by Adagrad in Algorithm \ref{alg:adagrad}, where $g_t \in \partial f_t(\theta_t)$ 
and $f_t:C \rightarrow \R$ is an arbitrary convex function, then for stepsize $\alpha>0$
the regret is upper bounded as
\[ R(T) \leq \frac{D_\infty^2}{2\alpha} \sum_{i=1}^d \norm{g_{1:T,i}}_2 + \alpha \sum_{i=1}^d \norm{g_{1:T,i}}_2.\]
\end{theorem}
The effective step-length of Adagrad is on the order of $\frac{\alpha}{\sqrt{t}}$. This can be seen as follows; first note that $v_{T,i} = \sum_{t=1}^T g^2_{t,i}$ and thus $(A_t)^{-1}$ is a diagonal matrix with entries $\frac{1}{\sqrt{v_{t,i}}+\delta}$. Then one has
\begin{align}\label{eq:adagrad}
      \alpha (A_T^{-1})_{ii} &= \frac{\alpha}{\sqrt{\sum_{t=1}^T g^2_{t,i}}+\delta} \nonumber\\&= \frac{\alpha}{\sqrt{T}}\frac{1}{\sqrt{\frac{1}{T}\sum_{t=1}^T g^2_{t,i}}+\frac{\delta}{\sqrt{T}}}
\end{align}
Thus an alternative point of view of Adagrad, is that it has a decaying stepsize $\frac{\alpha}{\sqrt{t}}$ but now  the correction term
becomes the running average of the squared derivatives  plus a vanishing damping term. However,  the effective stepsize has to decay
faster to get a logarithmic regret bound for the strongly convex case. This is what we analyze in the next section, where we propose
SC-Adagrad for strongly convex functions. 

\section{Strongly convex Adagrad (SC-Adagrad)}
The modification SC-Adagrad of Adagrad which we propose in the following can be motivated by the observation that the online projected gradient descent  \cite{hazan2007logarithmic} uses stepsizes of order $\alpha = O(\frac{1}{T})$ in order to achieve the logarithmic regret bound for strongly convex functions. In analogy with the derivation in the previous section, we still have $v_{T,i} = \sum_{t=1}^T g^2_{t,i}$. But now we modify  $(A_t)^{-1}$ 
and set it as a diagonal matrix with entries $\frac{1}{v_{t,i}+\delta_t}$. Then one has
\begin{align}\label{eq:sc-adagrad}
      \alpha (A_T^{-1})_{ii} = \frac{\alpha}{\sum_{t=1}^T g^2_{t,i}+\delta_t} = \frac{\alpha}{T}\frac{1}{\frac{1}{T}\sum_{t=1}^T g^2_{t,i}+\frac{\delta_T}{T}}.
\end{align}
Again, we have in the denominator a running average of the observed gradients and a decaying damping factor.
In this way, we get an effective stepsize of order $O(\frac{1}{T})$ in SC-Adagrad. The formal method is presented in Algorithm \ref{alg:sc_adagrad}.
As just derived the only difference of Adagrad and SC-Adagrad is the definition of the diagonal matrix $A_t$. 
\begin{algorithm}[ht]
   \caption{SC-Adagrad}
   \label{alg:sc_adagrad}
\begin{algorithmic}
   \STATE {\bfseries Input:} $\theta_1 \in C$\,,\,$\delta_0 > 0 , v_0 = 0 \in \mathbb{R}^d$
   \FOR{$t=1$ {\bfseries to} T}
   \STATE $g_t \in \partial f_t(\theta_t)$
   \STATE $v_t = v_{t-1}+(g_t \odot g_t)$
   \STATE Choose $0 < \delta_t \leq \delta_{t-1}$ element wise
   \STATE $A_t = \diag(v_t)+\diag(\delta_t)$
   \STATE $\theta_{t+1} = P_C^{A_t}\big(\theta_t - \alpha A_t^{-1}g_t\big)$
   \ENDFOR
\end{algorithmic}
\end{algorithm}
Note also that we have defined the damping factor $\delta_t$ as a function of $t$ which is also different from standard Adagrad. The constant $\delta$ in Adagrad is mainly introduced due to numerical reasons in order to avoid problems when $g_{t,i}$ is very small for some components in the first iterations and 
is typically chosen quite small e.g. $\delta=10^{-8}$. For SC-Adagrad the situation is different. If the first components $g_{1,i}, g_{2,i},\ldots$ are very small, say of order $\epsilon$, then the update is $\frac{\eps}{\eps^2 + \delta_t}$ which can become extremely large if $\delta_t$ is chosen to be small.
This would make the method very unstable and would lead to huge constants in the bounds. This is probably why in \cite{Rud2016}, the  modification
of Adagrad where one ``drops the square-root'' did not work. A good choice of  $\delta_t$ should be initially roughly on the order of $1$ and it should decay as $v_{t,i}=\sum_{t=1}^T g^2_{t,i}$ starts to grow. This is why we propose to use 
\[ \delta_{t,i}= \xi_2e^{-\xi_1 v_{t,i}}, \quad i=1,\ldots,d,\]
 for $\xi_1>0$, $\xi_2 > 0$ as a potential decay scheme as it satisfies both properties for sufficiently large $\xi_1$  and $\xi_2$ chosen on the order of 1. Also, one can achieve a constant decay scheme for $\xi_1 = 0\,,\,\xi_2 > 0$. We will come back to this choice after the proof.
In the following we provide the regret analysis of SC-Adagrad and show that the optimal logarithmic regret bound can be achieved. However, as it is data-dependent it is typically significantly better in practice than data-independent bounds.

\textbf{Note:} There was initial work in  \cite{duchi2010adaptive}  where SC-Adagrad using  a constant decay scheme was already proposed and also shown to have $O(\log T)$ regret bounds for strongly convex problems. Unfortunately, we were unaware of this work when we created SC-Adagrad, as it wasn't mentioned in journal version \cite{duchi2011adaptive}. Recently, there was a related work done for constant decay scheme in \cite{gupta2017unified} which gives a unified analysis of all the adaptive algorithms and some of the results regarding logarithmic regret bounds (see for Section 4 in \cite{gupta2017unified}) closely match to that of SC-Adagrad (Algorithm \ref{alg:sc_adagrad}). Our new contribution is that the decay scheme is vectorized so one need not restrict to a constant scheme. We only require a mild condition, that it is non-increasing element-wise in order to achieve logarithmic regret bounds.
\subsection{Analysis}
For any two matrices $A,B \in \mathbb{R}^{d \times d}$, we use the notation $\bullet$ to denote the inner product  i.e $A\bullet B = \sum_{i}\sum_j A_{ij}B_{ij}$. Note that $A \bullet B = \tr(A^T B)$. 
\begin{lemma}\label{lem:psd_lem} [Lemma 12 \cite{hazan2007logarithmic}]
Let $A,B$ be positive definite matrices, let $A \succeq B \succ 0$ then
\begin{align}
A^{-1}\bullet (A-B) \leq \log\Big(\frac{|A|}{|B|}\Big)
\end{align}
where $|A|$ denotes the determinant of the matrix $A$
\end{lemma}
\begin{lemma}\label{lem:sc-lem}
Let Assumptions \textbf{A1}, \textbf{A2} hold, then for $T\geq 1$ and $A_t,\delta_t$ as defined in the SC-Adagrad algorithm we have,
\begin{align*}
 \sum_{t=1}^{T}\inner{g_t, A_t^{-1}g_t} &\leq \sum_{i=1}^{d}\log\Bigg(\frac{\|g_{1:T,i}\|^2+\delta_{T,i}}{\delta_{1,i}}\Bigg)\\ & - \sum_{i=1}^d\sum_{t=2}^T\frac{\delta_{t,i}-\delta_{t-1,i}}{\|g_{1:t,i}\|^2 + \delta_{t,i}}
\end{align*} 
\end{lemma}
\ifpaper
\begin{proof}
Consider the following summation,
\begin{align*}
& \sum_{t=1}^{T}\inner{g_t, A_t^{-1}g_t} \leq \sum_{t=1}^{T} A_{t}^{-1} \bullet  \diag(g_tg_t^T)\\
&= A_{1}^{-1} \bullet (A_1 - \diag(\delta_1))\\
&+\sum_{t=2}^{T} A_{t}^{-1} \bullet  (A_t -A_{t-1} - \diag(\delta_t) + \diag(\delta_{t-1}) )\\
&\leq  \log\Big(\frac{|A_1|}{|\diag(\delta_1)|} \Big)+\sum_{t = 2}^T \log\Big(\frac{|A_t|}{|A_{t-1}|} \Big)\\
& - \sum_{t=2}^TA_{t}^{-1} \bullet (\diag(\delta_t) - \diag(\delta_{t-1})) \\
& = \log\Big(\frac{|A_T|}{|\diag(\delta_1)|}\Big) - \sum_{t=2}^TA_{t}^{-1} \bullet (\diag(\delta_t) - \diag(\delta_{t-1}))\\
&\leq \sum_{i=1}^{d}\log\Bigg(\frac{\|g_{1:T,i}\|^2+\delta_{T,i}}{\delta_{1,i}}\Bigg) \\
&- \sum_{t=2}^TA_{t}^{-1} \bullet (\diag(\delta_t) - \diag(\delta_{t-1}))\\ 
&\leq \sum_{i=1}^{d}\log\Bigg(\frac{\|g_{1:T,i}\|^2+\delta_{T,i}}{\delta_{1,i}}\Bigg)  - \sum_{i=1}^d\sum_{t=2}^T\frac{\delta_{t,i}-\delta_{t-1,i}}{\|g_{1:t,i}\|^2 + \delta_{t,i}}
\end{align*}
In the first step we use $\inner{x,Ax}  = A \bullet \diag(xx^T)$ where $A$ is a diagonal matrix and subsequently we use\,$\forall t>1$\,, $\diag(g_tg_t^T) = A_t - A_{t-1} - \diag(\delta_t) + \diag(\delta_{t-1})$, and for\,\,$t=1$\, we have $diag(g_1g_1^T) = A_1 - diag(\delta_1)$. In the first inequality we use Lemma \ref{lem:psd_lem} also see for Lemma 12 of \cite{hazan2007logarithmic}. 
Note that for $T=1$, the upper bound results in 0.
\end{proof}
\fi
\begin{theorem}\label{thm:sc-thm}
Let Assumptions \textbf{A1}, \textbf{A2} hold and let ${\theta_t}$ be the sequence generated by the SC-Adagrad in Algorithm \ref{alg:sc_adagrad}, where $g_t \in \partial f_t(\theta_t)$  and $f_t:C \rightarrow \R$ is an arbitrary $\mu$-strongly convex function $(\mu \in \R^d_+)$ where the stepsize fulfills $\alpha \geq \max_{i=1,\ldots,d} \frac{G_{\infty}^2}{2\mu_i}$.
Furthermore, let $\delta_t > 0$  and $\delta_{t,i} \leq \delta_{t-1,i} \forall t \in [T], \forall i \in [d]$, then
the regret of SC-Adagrad can be upper bounded for $T\geq 1$ as
\begin{align*}
 & R(T) \leq \frac{D_{\infty}^2\tr(\diag(\delta_1))}{2\alpha} + \frac{\alpha}{2}\sum_{i=1}^{d}\log\Big(\frac{\|g_{1:T,i}\|^2+\delta_{T,i}}{\delta_{1,i}}\Big)\\
 & + \frac{1}{2}\sum_{i=1}^{d} \inf_{t \in [T]}\Big(\frac{(\theta_{t,i}-\theta_{i}^*)^2}{\alpha}- \frac{\alpha}{\|g_{1:t,i}\|^2 + \delta_{t,i}}\Big) (\delta_{T,i}-\delta_{1,i})
\end{align*}
For constant $\delta_t$ i.e $\delta_{t,i} = \delta > 0\, \forall t \in [T] \,\text{and}\, \forall i \in [d]$ then the regret of SC-Adagrad is upper bounded as 
\begin{align}
\label{bound:sc_adagrad2}
& R(T) \leq \frac{D_{\infty}^2d\delta}{2\alpha} + \frac{\alpha}{2}\sum_{i=1}^{d}\log\Big(\frac{\|g_{1:T,i}\|^2+\delta}{\delta}\Big)
\end{align}
For $\zeta$-strongly convex function choosing $\alpha \geq \frac{G_{\infty}^2}{2\zeta}$ we obtain the above mentioned regret bounds.
\end{theorem}
\ifpaper
\begin{proof}
We rewrite the regret bound with the definition of $\mathbf{\mu}$-strongly convex functions as 
\begin{align*}
& R(T)=\sum_{t=1}^T (f_t(\theta_t)-f_t(\theta^*)) \\
& \leq \sum_{t=1}^T \inner{g_t,\theta_{t}-\theta^*} - \sum_{t=1}^T \norm{\theta_t-\theta^*}^2_{\diag(\mu)}
\end{align*}
Using the non-expansiveness we have
\begin{align*}
& \norm{\theta_{t+1}-\theta^*}^2_{A_t} \\
&= \norm{P_C^{A_t}\big(\theta_t-\alpha A^{-1}_t g_t\big)-\theta^*}^2_{A_t}\\
&\leq  \norm{\theta_t-\alpha A_t^{-1} g_t-\theta^*}^2_{A_t}\\
&\leq \|\theta_t - \theta^*\|^2_{A_t} - 2\alpha\inner{g_t, \theta_t - \theta^*}+ \alpha^2\inner{g_t,A_t^{-1}g_t}
\end{align*}
This yields
\begin{align*}
& \inner{g_t,\theta_t - \theta^*}\\
 &\leq \frac{\|\theta_t-\theta^*\|^2_{A_t} - \|\theta_{t+1}-\theta^*\|^2_{A_t}}{2\alpha} + \frac{\alpha}{2}\inner{g_t,A_t^{-1}g_t}
\end{align*}
Hence we can upper bound the regret as follows
\begin{align*}
& R(T)\\
&\leq \sum_{t=1}^{T}\frac{\|\theta_t-\theta^*\|^2_{A_t} - \|\theta_{t+1}-\theta^*\|^2_{A_t}}{2\alpha} \\
& + \frac{\alpha}{2}\sum_{t=1}^{T}\inner{g_t,A_t^{-1}g_t} - \sum_{t=1}^T \norm{\theta_t-\theta^*}^2_{\diag(\mu)}\\
&\leq \frac{\|\theta_1 - \theta^*\|^2_{A_1}}{2\alpha} +\sum_{t=2}^{T}\Big(\frac{\|\theta_t-\theta^*\|^2_{A_t} - \|\theta_{t}-\theta^*\|^2_{A_{t-1}}}{2\alpha}\Big)  \\
&- \frac{\|\theta_{T+1}-\theta^*\|^2_{A_{T}}}{2\alpha} + \frac{\alpha}{2}\sum_{t=1}^{T}\inner{g_t,A_t^{-1}g_t}\\
&- \sum_{t=1}^T \norm{\theta_t-\theta^*}^2_{\diag(\mu)}\\
&\leq \frac{\|\theta_1 - \theta^*\|^2_{A_1-2\alpha \diag(\mu)}}{2\alpha}\\
& +\sum_{t=2}^{T}\frac{\|\theta_t-\theta^*\|^2_{A_t - A_{t-1} -2\alpha\, \diag(\mu)} }{2\alpha}+ \frac{\alpha}{2}\sum_{t=1}^{T}\inner{g_t, A_t^{-1}g_t}
\end{align*}
In the last step we use the equality $\forall x \in \mathbb{R}^n$ $\|x\|^2_A - \|x\|^2_B  = \|x\|^2_{A-B}$ where  $A,B \in \mathbb{R}^{n\,\text{x}\,n}$ and both are diagonal matrices. Now, we choose $\alpha$ such that $A_t - A_{t-1} - 2\alpha\, \diag(\mu) \preccurlyeq \diag(\delta_t) - \diag(\delta_{t-1}) $ $\forall t\geq 2$ and $A_1 - 2\alpha\, \diag(\mu) \preccurlyeq \diag(\delta_1) $  Since $A_t - A_{t-1}  \preccurlyeq G_{\infty}^2I + \diag(\delta_t) - \diag(\delta_{t-1})$ and $A_1 \preccurlyeq G_{\infty}^2I +\diag(\delta_1)$\, because at any round the difference between subsequent squares of sub-gradients is bounded by $G_{\infty}^2$. Also by Algorithm \ref{alg:sc_adagrad}, $\delta_{t,i} \leq \delta_{t-1,i} \forall t>1 , \forall i \in [d]$ hence $\diag(\delta_t) - \diag(\delta_{t-1}) \preceq 0$. Hence by choosing $\alpha \geq \max_{i=1,\ldots,d} \frac{ G^2_{\infty}}{2\mu_i}$  we have $A_t - A_{t-1} - 2\alpha\, \diag(\mu) \preccurlyeq \diag(\delta_t) - \diag(\delta_{t-1})  \, \forall t \geq 2$ and $A_1 - 2\alpha \diag(\mu) \preccurlyeq \diag(\delta_1)$ which yields
\begin{align*}
&R(T) \\
&\leq\frac{\|\theta_1 - \theta^*\|^2_{\diag(\delta_1)}}{2\alpha}+ \sum_{t=2}^{T}\frac{\|\theta_t - \theta^*\|_{\diag(\delta_t)-\diag(\delta_{t-1})}}{2\alpha}\\
&+ \frac{\alpha}{2}\sum_{t=1}^{T}\inner{g_t, A_t^{-1}g_t}\\
&= \frac{\|\theta_1 - \theta^*\|^2_{\diag(\delta_1)}}{2\alpha}+ \sum_{t=2}^{T}\sum_{i=1}^{d}\frac{(\theta_{t,i}-\theta_{i}^*)^2(\delta_{t,i}-\delta_{t-1,i})}{2\alpha}\\
& + \frac{\alpha}{2}\sum_{t=1}^{T}\inner{g_t, A_t^{-1}g_t}\\
&\leq\frac{D_{\infty}^2\tr(\diag(\delta_1))}{2\alpha}+ \frac{\alpha}{2}\sum_{t=1}^{T}\inner{g_t, A_t^{-1}g_t}\\
& + \sum_{t=2}^{T}\sum_{i=1}^{d}\frac{(\theta_{t,i}-\theta_{i}^*)^2(\delta_{t,i}-\delta_{t-1,i})}{2\alpha}\\
&\leq \frac{D_{\infty}^2\tr(\diag(\delta_1))}{2\alpha} + \frac{\alpha}{2}\sum_{i=1}^{d}\text{log}\Big(\frac{\|g_{1:T,i}\|^2+\delta_{T,i}}{\delta_{1,i}}\Big) \\
& + \frac{1}{2}\sum_{t=2}^{T}\sum_{i=1}^{d}\Bigg(\frac{(\theta_{t,i}-\theta_{i}^*)^2(\delta_{t,i}-\delta_{t-1,i})}{\alpha}- \frac{\alpha(\delta_{t,i}-\delta_{t-1,i})}{\|g_{1:t,i}\|^2 + \delta_{t,i}}\Bigg)\\
&\leq \frac{D_{\infty}^2\tr(\diag(\delta_1))}{2\alpha} + \frac{\alpha}{2}\sum_{i=1}^{d}\text{log}\Big(\frac{\|g_{1:T,i}\|^2+\delta_{T,i}}{\delta_{1,i}}\Big) \\
& + \frac{1}{2}\sum_{i=1}^{d} \inf_{t \in [T]}\Big(\frac{(\theta_{t,i}-\theta_{i}^*)^2}{\alpha}- \frac{\alpha}{\|g_{1:t,i}\|^2 + \delta_{t,i}}\Big) (\delta_{T,i}-\delta_{1,i})
\end{align*}
In the second inequality we bounded $\|\theta_1 - \theta^*\|_{\diag(\delta_1)}^2 \leq D_{\infty}^2\tr(\diag(\delta_1))$. In the second last step we use the Lemma \ref{lem:sc-lem}. So under a constant $\delta_t$ i.e $\delta_{t,i} = \delta > 0, \, \forall t \in [T], \forall i \in [d]$ we have $\tr(\diag(\delta_1)) = d\delta$ hence proving the result \eqref{bound:sc_adagrad2}. For $\zeta$-strongly convex functions choosing $\alpha \geq \frac{G_{\infty,i}^2}{2\zeta}$ we obtain the the same results as $\mu$-strongly convex functions. This can be seen by setting $\mu_i = \zeta,\; \forall i \in [d]$.
\end{proof}
\fi
Note that the first and the last term in the regret bound can be upper bounded by constants. Only the second term depends 
on $T$. Note that $\norm{g_{1:T,i}}^2 \leq T G^2$ and as $\delta_t$ is monotonically decreasing, the second term is 
on the order of $O(\log(T))$ and thus we have a logarithmic regret bound. As the bound is data-dependent, in the sense
that it depends on the observed sequence of gradients, it is much tighter than a data-independent bound.

The bound includes also the case of a non-decaying damping factor $\delta_t=\delta=\xi_2$ ($\xi_1=0$). While a rather large constant 
damping factor can work well, we have noticed that the best results are obtained with the decay scheme
\[ \delta_{t,i}= \xi_2e^{-\xi_1 v_{t,i}}, \quad i=1,\ldots,d.\]
where $\xi_1> 0\,,\,\xi_2 > 0\,$, which is what we use in the experiments. Note that this decay scheme for $\xi_1, \xi_2 > 0$ is adaptive to the specific dimension and thus increases the adaptivity of the overall algorithm. For completeness we also give the bound specialized for this decay scheme.
\begin{corollary}\label{cor:sc-thm}
In the setting of Theorem \ref{thm:sc-thm} choose $\delta_{t,i}= \xi_2e^{-\xi_1 v_{t,i}}$ for $i=1,\ldots, d$ for some $\xi_1 > 0, \xi_2 > 0\,$. Then the regret of SC-Adagrad  can be upper bounded for $T\geq 1$ as
\begin{align*}
 & R(T) \leq \frac{d D_{\infty}^2\xi_2}{2\alpha} -\frac{\alpha}{2}\log(\xi_2e^{-\xi_1 G_{\infty}^2})\\
 &+ \frac{\alpha }{2}\sum_{i=1}^{d}\log\Big(\|g_{1:T,i}\|^2+\xi_2\,e^{-\xi_1\|g_{1:T,i}\|^2}\Big)\\
 & + \frac{\alpha\xi_1\xi_2 }{2\big(\log(\xi_2\,\xi_1)+1\big)}\sum_{i=1}^d \Big(1-e^{-\xi_1\|g_{1:T,i}\|^2}\Big)
\end{align*}
\end{corollary}
\ifpaper
\begin{proof}
Note that $\delta_{T,i}=\xi_2e^{-\xi_1 v_{T,i}} = \xi_2e^{-\xi_1 \norm{g_{1:T,i}}^2}$.
Plugging this into Theorem \ref{cor:sc-thm} for $\xi_1,\xi_2 > 0$ yields the results for the first three terms.  Using $(\theta_{t,i}-\theta_{i}^*)^2 \geq 0$ we have 
\begin{align*}
&\inf_{t \in [T]}\Big(\frac{(\theta_{t,i}-\theta_{i}^*)^2}{\alpha}- \frac{\alpha}{\|g_{1:t,i}\|^2 + \delta_{t,i}}\Big) \\
&\geq \frac{-\alpha}{\inf_{j\in [1:T]}\|g_{1:j,i}\|^2 + \delta_{j,i}}
\end{align*} 
Note that $\norm{g_{1:j,i}}^2+\delta_{j,i} = v_{j,i} + \xi_2e^{-\xi_1 v_{j,i}}$, in order to find the minimum of this term we thus analyze the function $f:\R_+ \rightarrow \R$, $f(x)=x+\xi_2e^{-\xi_1 x}$. 
and a straightforward calculation shows that the minimum is attained at $x^*=\frac{1}{\xi_1}\log(\xi_1\xi_2)$ and $f(x^*)=\frac{1}{\xi_1}(\log(\xi_1\xi_2)+1)$. This yields the fourth term.
\end{proof}
\fi
Unfortunately, it is not obvious that the regret bound for our decaying damping factor is better than the one of a constant damping factor.
Note, however that the third term in the regret bound of Theorem \ref{thm:sc-thm} can be negative. It thus remains an interesting question for future work, if there exists an optimal decay scheme which provably works better than any
constant one.

\section{RMSProp and SC-RMSProp}


RMSProp is one of the most popular adaptive gradient algorithms used for the training of deep neural networks
\cite{Sch2014,Dau2015,Dan2016,Schmidhuber201585}. It has been used frequently in computer vision \cite{Kar2016} e.g. to train the latest InceptionV4 network \cite{SzeEtAl2016,SzeEtAl2016b}. Note that RMSProp outperformed other adaptive methods like Adagrad order Adadelta as well as SGD with momentum in a large number of tests in \cite{Sch2014}. It has been argued that if the changes in the parameter update are approximately Gaussian distributed, then the matrix $A_t$ can be seen as a preconditioner which approximates the diagonal of the Hessian \cite{Dan2016}. However, it is fair to say that despite its huge empirical success in practice and some first analysis in the literature, there is so far no rigorous theoretical analysis of RMSProp. We will 
analyze RMSProp given in Algorithm \ref{alg:rmsprop} in the framework of of online convex optimization.

\begin{algorithm}[ht]
   \caption{RMSProp}
   \label{alg:rmsprop}
\begin{algorithmic}
   \STATE {\bfseries Input:} $\theta_1 \in C$\,,\,$\delta>0,\, \alpha > 0 , v_0 = 0 \in \mathbb{R}^d$
   \FOR{$t=1$ {\bfseries to} T}
   \STATE $g_t \in \partial f_t(\theta_t)$
   \STATE $v_t = \beta_tv_{t-1}+(1-\beta_t)(g_t \odot g_t)$
   \STATE Set $\epsilon_t = \frac{\delta}{\sqrt{t}}$ and $\alpha_t = \frac{\alpha}{\sqrt{t}}$
   \STATE $A_t = diag(\sqrt{v_t})+\eps_t I$\;
   \STATE $\theta_{t+1} = P_C^{A_t}\big(\theta_t - \alpha_t A_t^{-1}g_t\big)$
   \ENDFOR
\end{algorithmic}
\end{algorithm}
First, we will show that RMSProp reduces to Adagrad for a certain choice of its parameters. Second, we will prove for the general convex case a regret bound of  $O(\sqrt{T})$ similar to the bound given in Theorem \ref{th:adagrad}. It turns out
that the convergence analysis requires that in the update of the weighted cumulative squared gradients ($v_t$)\,, it has to hold
\[ 1-\frac{1}{t}\, \leq \, \beta_{t} \, \leq 1-\frac{\gamma}{t},\]
for some $0<\gamma\leq 1$. This is in contrast to the original suggestion of \cite{hinton2012lecture} to choose $\beta_t=0.9$.
It will turn out later in the experiments that the constant choice of $\beta_t$ leads sometimes to divergence of the sequence, whereas the choice derived from our theoretical analysis always leads to a convergent scheme even when applied to
deep neural networks. Thus we think that the analysis in the following is not only interesting for the convex case but can give
valuable hints how the parameters of RMSProp should be chosen in deep learning.

Before we start the regret analysis we want to discuss the sequence $v_t$ in more detail. Using the recursive definition
of $v_t$, we get the closed form expression
\[ v_{t,i} = \sum_{j=1}^t  (1-\beta_j) \prod_{k=j+1}^t \beta_k g^2_{j,i}.\]
With $\beta_t =  1 - \frac{1}{t}$ one gets, $v_{t,i} = \sum_{j=1}^t \frac{1}{j} \prod_{k=j+1}^t \frac{k-1}{k} g^2_{j,i}$,
and using the telescoping product ones gets $\prod_{k=j+1}^t \frac{k-1}{k}=\frac{j}{t}$ and thus

 $v_{t,i}=\frac{1}{t}\sum_{j=1}^t g^2_{j,i}$.

If one uses additionally the stepsize scheme $\alpha_t = \frac{\alpha}{\sqrt{t}}$ and $\eps_t=\frac{\delta}{\sqrt{T}}$,
then we recover the update scheme of Adagrad, see \eqref{eq:adagrad}, as a particular case of RMSProp. We are not
aware of that this correspondence of Adagrad and RMSProp has been observed before.

 
The proof of the regret bound for RMSProp relies on the following lemma.
\begin{lemma}\label{lem:rms_lem}
Let Assumptions \textbf{A1} and \textbf{A2} and suppose that  $1-\frac{1}{t}\, \leq \, \beta_{t} \, \leq 1-\frac{\gamma}{t}$  for some $0<\gamma\leq 1$,  and $t\geq 1$. Also for $t>1$ suppose $\sqrt{(t-1)}\eps_{t-1} \leq \sqrt{t}\eps_t$, then
\[ \sum_{t=1}^T \frac{g^2_{t,i}}{\sqrt{t\,v_{t,i} }+\sqrt{ t}\eps_t} \, \leq \, \frac{2(2-\gamma)}{\gamma}\Big(\sqrt{T\,v_{T,i} }+ \sqrt{T}\eps_T\Big).\]
\end{lemma}
\ifpaper
\begin{proof}
The lemma is proven via induction. For $T=1$ we have $v_0=0$ and thus $v_{1,i}=(1-\beta_{1})g^2_{1,i}$ and thus
\begin{align*}
 & \frac{g^2_{1,i}}{(\sqrt{v_{1,i}}+\eps_1)} = \frac{(1-\beta_{1})g^2_{t,i}}{(1-\beta_{1})\Big(\sqrt{(1-\beta_{1})g^2_{1,i} }+ \eps_1\Big)} \\
 &\leq \, \frac{\sqrt{(1-\beta_{1})g^2_{1,i} }+ \eps_1}{1-\beta_{1}} \leq \frac{(\sqrt{v_{1,i}}+\eps_1)}{\gamma}.
\end{align*}

Note that  $\frac{1}{\gamma}\leq \frac{2(2-\gamma)}{\gamma}$ since $2(2-\gamma)>1$ for $\gamma \leq 1$ hence the bound holds for $T=1$. For $T>1$ we suppose that the bound is true for $T-1$ and get
\begin{align*}
 &\sum_{t=1}^{T-1}\frac{g^2_{t,i}}{\sqrt{t\,v_{t,i} }+ \sqrt{t}\eps_t} \\
 &\, \leq \, \frac{2(2-\gamma)}{\gamma} \Big(\sqrt{(T-1)\,v_{T-1,i} }+ \sqrt{(T-1)}\eps_{T-1}\Big).
\end{align*}
We rewrite $v_{T,i} = \beta_{T}\,v_{T-1,i} + (1-\beta_{T})g^2_{T,i}$ as $v_{T-1,i} = \frac{1}{\beta_{T}} v_{T,i} - \frac{1-\beta_{T}}{\beta_{T}}g^2_{T,i}$ and with $ \sqrt{(t-1)}\eps_{t-1} \leq \sqrt{t}\eps_t $ we get
\begin{align*}
&\sqrt{(T-1)v_{T-1,i}}+\sqrt{(T-1)}\eps_{T-1}\\
\leq&\sqrt{\frac{T-1}{\beta_{T}} v_{T,i} - \frac{(T-1)(1-\beta_{T})}{\beta_{T}}g^2_{T,i}}+\sqrt{T}\eps_{T}\\
=&\sqrt{\frac{T-1}{T\beta_{T}} Tv_{T,i} - \frac{(T-1)(1-\beta_{T})}{\beta_{T}}g^2_{T,i}}+\sqrt{T}\eps_{T}\\
\leq &\sqrt{Tv_{T,i} - \frac{(T-1)(1-\beta_{T})}{\beta_{T}}g^2_{T,i}}+\sqrt{T}\eps_{T}\\
\end{align*}
Note that in the last step we have used that $\frac{T-1}{T\beta_{T}} \leq 1$ and the fact that $\sqrt{x}$  is concave and thus $\sqrt{x-c} \leq \sqrt{x} - \frac{c}{2\sqrt{x}}$ given that $x-c\geq 0$, along with $\frac{-1}{\sqrt{T v_{T,i}}}\leq \frac{-1}{\sqrt{T v_{T,i}}+\sqrt{T}\eps_t}$ for $v_{t,i} \neq 0$ we have 
\begin{align*}
& \sqrt{(T-1)v_{T-1,i}}+\sqrt{(T-1)}\eps_{T-1} \\
& \leq\sqrt{Tv_{T,i}}+\sqrt{T}\eps_{T} -  \frac{(T-1)(1-\beta_{T})}{2\beta_{T}\big(\sqrt{Tv_{T,i}}+\sqrt{T}\eps_{T}\big)}g^2_{T,i} \numberthis \label{eqn:eq1}
\end{align*}
Using the above bound we have the following
\begin{align*}
&\sum_{t=1}^{T}\frac{g^2_{t,i}}{\sqrt{t\,v_{t,i}}+\sqrt{t}\eps_t}\\
&= \sum_{t=1}^{T-1}\frac{g^2_{t,i}}{\sqrt{t\,v_{t,i}}+\sqrt{t}\eps_t} + \frac{g_{T,i}^2}{\sqrt{Tv_{T,i}} + \sqrt{T}\epsilon_T} \\
& \leq \frac{2(2-\gamma)}{\gamma} \Big(\sqrt{(T-1)\,v_{T-1,i} }+ \sqrt{(T-1)}\eps_{T-1}\Big)\\
& + \frac{g^2_{T,i}}{\sqrt{T\,v_{T,i}}+\sqrt{T}\eps_t}\\
 &  \leq \frac{2(2-\gamma)}{\gamma}\Big(\sqrt{Tv_{T,i}}+\sqrt{T}\eps_{T}\Big)\\
& +\Big( 1 -   \frac{2(2-\gamma)}{\gamma}\frac{(T-1)(1-\beta_{T})}{2\beta_{T}}\Big)\frac{g^2_{T,i}}{\sqrt{Tv_{T,i}}+\sqrt{T}\eps_{T}}
\end{align*}
In the last step we use \eqref{eqn:eq1} and since for $T > 1$ the term $\Big( 1 -   \frac{2(2-\gamma)}{\gamma}\frac{(T-1)(1-\beta_{T})}{2\beta_{T}}\Big) \leq 0$ for $1-\frac{1}{t}\leq \beta_{t}\leq 1-\frac{\gamma}{t}$
\end{proof}
\fi

\begin{corollary}\label{cor:rms_cor}
Let Assumptions \textbf{A1}, \textbf{A2} hold and suppose that  $1-\frac{1}{t}\, \leq \, \beta_{t} \, \leq 1-\frac{\gamma}{t}$  for some $0<\gamma\leq 1$,  and $t\geq 1$. Also for $t>1$ suppose $\sqrt{(t-1)}\eps_{t-1} \leq \sqrt{t}\eps_t$, and set $\alpha_t=\frac{\alpha}{\sqrt{t}}$, then
\begin{align*}
\sum_{t=1}^{T}\frac{\alpha_t}{2}\inner{g_t, A_t^{-1}g_t} \leq 
\frac{\alpha (2-\gamma)}{\gamma}\sum_{i=1}^d \Big(\sqrt{T\,v_{T,i} }+ \sqrt{T}\eps_T\Big)
\end{align*}
\end{corollary}
\ifpaper
\begin{proof}
Using the definition of $A_t = diag(\sqrt{v_t}) + \epsilon_t I$, $\alpha_t = \frac{\alpha}{\sqrt{t}}$ along with Lemma \ref{lem:rms_lem} we get
\begin{align*}
\sum_{t=1}^{T}\frac{\alpha_t}{2}\inner{g_t, A_t^{-1}g_t} &=  \sum_{t=1}^T \frac{\alpha_t}{2}\sum_{i=1}^d \frac{g_{t,i}^2}{\sqrt{v_{t,i} }+\eps_t}\\
&= \frac{\alpha}{2} \sum_{t=1}^T \sum_{i=1}^d \frac{g_{t,i}^2}{\sqrt{t\,v_{t,i} }+\sqrt{t}\eps_t}\\
&\leq \frac{\alpha}{2}\sum_{i=1}^d \frac{2(2-\gamma)}{\gamma}\Big(\sqrt{T\,v_{T,i} }+ \sqrt{T}\eps_T\Big)
\end{align*}
\end{proof}
\fi
With the help of  Lemma \ref{lem:rms_lem} and Corollary \ref{cor:rms_cor} we can now state the regret bound for RMSProp.
\begin{theorem}\label{thm:temp_th1}
Let Assumptions \textbf{A1}, \textbf{A2} hold and let ${\theta_t}$ be the sequence generated by RMSProp in Algorithm \ref{alg:rmsprop}, where $g_t \in \partial f_t(\theta_t)$  and $f_t:C \rightarrow \R$ is an arbitrary convex function
and $\alpha_t=\frac{\alpha}{\sqrt{t}}$ for some $\alpha>0$ and $1-\frac{1}{t}\, \leq \, \beta_{t} \, \leq 1-\frac{\gamma}{t}$  for some $0<\gamma\leq 1$. Also for $t>1$ let $\sqrt{(t-1)}\eps_{t-1} \leq \sqrt{t}\eps_t$, then the regret of RMSProp can be upper bounded for $T\geq 1$ as
\[
R(T) \leq  \Big(\frac{D_\infty^2}{2\alpha}+ \frac{\alpha (2-\gamma)}{\gamma}\Big) \sum_{i=1}^d  \Big(\sqrt{Tv_{T,i}} +\sqrt{T}\eps_T\Big)
\]
\end{theorem}
\ifpaper
\begin{proof}
Note that for every convex function $f_t:C \rightarrow \R$ it holds for all $x,y \in C$ and $g_t \in \partial f_t(x)$,
\[ f_t(y) \geq f_t(x) + \inner{g_t, y-x}.\]
We use this to upper bound the regret as
\begin{align*}
R(T)=\sum_{t=1}^T (f_t(\theta_t)-f_t(\theta^*)) \leq \sum_{t=1}^T \inner{g_t,\theta_{t}-\theta^*} 
\end{align*}
Using the non-expansiveness of the weighted projection, we have
\begin{align*}
& \norm{\theta_{t+1}-\theta^*}^2_{A_t} \\
&= \norm{P_C^{A_t}\big(\theta_t-\alpha_t A_t^{-1} g_t\big)-\theta^*}^2_{A_t}\\
&\leq  \norm{\theta_t-\alpha_t A_t^{-1} g_t-\theta^*}^2_{A_t}\\
&\leq \|\theta_t - \theta^*\|^2_{A_t} - 2\alpha_t\inner{g_t, \theta_t - \theta^*}+ \alpha_t^2\inner{g_t,A_t^{-1}g_t}
\end{align*}
This yields
\begin{align*}
& \inner{g_t,\theta_t - \theta^*} \\
&\leq \frac{\|\theta_t-\theta^*\|^2_{A_t} - \|\theta_{t+1}-\theta^*\|^2_{A_t}}{2\alpha_t} + \frac{\alpha_t}{2}\inner{g_t,A_t^{-1}g_t}
\end{align*}
Hence we can upper bound the regret as follows
\begin{align*}
& R(T)\\
 &\leq \sum_{t=1}^{T}\frac{\|\theta_t-\theta^*\|^2_{A_t} - \|\theta_{t+1}-\theta^*\|^2_{A_t}}{2\alpha_t}\\
 & + \sum_{t=1}^{T}\frac{\alpha_t}{2}\inner{g_t,A_t^{-1}g_t}\\
&\leq\frac{\|\theta_1 - \theta^*\|^2_{A_1}}{2\alpha_1}+\sum_{t=2}^{T}\Big(\frac{\|\theta_t-\theta^*\|^2_{A_t}}{2\alpha_t} - \frac{\|\theta_t-\theta^*\|^2_{A_{t-1}}}{2\alpha_{t-1}}\Big)\\
& - \frac{\|\theta_{T+1}-\theta^*\|^2_{A_{T}}}{2\alpha_t}+ \sum_{t=1}^{T}\frac{\alpha_t}{2}\inner{g_t,A_t^{-1}g_t}\\
&\leq \frac{\|\theta_1 - \theta^*\|^2_{A_1}}{2\alpha} \\
&+ \sum_{t=2}^{T}\frac{\|\theta_t-\theta^*\|^2_{\sqrt{t}A_t - \sqrt{t-1}A_{t-1} } }{2\alpha} + \sum_{t=1}^{T}\frac{\alpha_t}{2}\inner{g_t, A_t^{-1}g_t}
\end{align*}
In the last step we used $\alpha_t = \frac{\alpha}{\sqrt{t}}$. We show that
\begin{equation}
\label{eqn:diag_eqn_cond}
\sqrt{t}A_t - \sqrt{t-1}A_{t-1} \succeq 0\, \forall t \in [T]
\end{equation}
Note that $A_t$ are diagonal matrices for all $t\geq 1$. We note that
\[ (A_t)_{ii} = \sqrt{v_{t,i}} + \epsilon_t,\]
and $v_{t,i}=\beta_t v_{t-1,i}+(1-\beta_t)g^2_{t,i}$ as well as $\beta_t \geq 1-\frac{1}{t}$ which implies $t\beta_t \geq t-1$.
We get
\begin{align*}
\sqrt{t}(A_t)_{ii} &= \sqrt{tv_{t,i}} + \sqrt{t}\epsilon_t \\
&=     \sqrt{t \beta_t v_{t-1,i} + t(1-\beta_t)g^2_{t,i}} + \sqrt{t}\epsilon_t\\
&\geq  \sqrt{(t-1) v_{t-1,i}} + \sqrt{t-1}\epsilon_{t-1},
\end{align*}
where we used in the last inequality that $\sqrt{t}\epsilon_t \geq \sqrt{t-1}\epsilon_{t-1}$.
Note that
\begin{align*}
   &\sum_{t=2}^{T} \|\theta_t-\theta^*\|^2_{\sqrt{t}A_t - \sqrt{t-1}A_{t-1} } \\
= &\sum_{t=2}^T \sum_{i=1}^d (\theta_{t,i}-\theta^*_i)^2 \\
   &\Big(\sqrt{tv_{t,i}} + \sqrt{t}\epsilon_t - \sqrt{(t-1) v_{t-1,i}} + \sqrt{t-1}\epsilon_{t-1}\Big)\\
\leq & \sum_{i=1}^d D_\infty^2\\
      & \sum_{t=2}^T \Big(\sqrt{tv_{t,i}} + \sqrt{t}\epsilon_t - \sqrt{(t-1) v_{t-1,i}} -\sqrt{t-1}\epsilon_{t-1}\Big)\\
= & \sum_{i=1}^d D_\infty^2 \Big(\sqrt{Tv_{T,i}} +\sqrt{T}\eps_T - \sqrt{v_{1,i}} - \eps_1\Big)
\end{align*}
where the inequality could be done as we showed before that the difference of the terms in $v_{t,i}$ is non-negative for all $i \in [d]$ and $t\geq 1$. As $(A_1)_{ii}=\sqrt{v_{1,i}}+\eps_1$ we get
\[  \frac{\|\theta_1 - \theta^*\|^2_{A_1}}{2\alpha} \leq \frac{D_\infty^2}{2\alpha} \sum_{i=1}^d \big(\sqrt{v_{1,i}} + \eps_1\big).\]
Thus in total we have
\begin{align*}
R(T) \leq &\frac{D_\infty^2 \sum_{i=1}^d  \Big(\sqrt{T v_{T,i}} +\sqrt{T}\eps_T\Big)}{2\alpha} \\&+ \sum_{t=1}^{T}\frac{\alpha_t}{2}\inner{g_t, A_t^{-1}g_t}.
\end{align*}
Finally, with Corollary \ref{cor:rms_cor} we get the result.
\end{proof}
\fi
Note that for $\beta_t=1-\frac{1}{t}$, that is $\gamma=1$, and $\eps_t=\frac{\delta}{T}$ where RMSProp corresponds to Adagrad  we recover the regret bound of Adagrad in the convex case, see Theorem \ref{th:adagrad}, up to the damping factor. Note that in this case
\[ \sqrt{T v_{T,i}} = \sqrt{\sum_{j=1}^T g^2_{j,i}} = \norm{g_{1:T,i}}_2.\]


\subsection{SC-RMSProp}
Similar to the extension of Adagrad to SC-Adagrad, we present in this section SC-RMSProp which achieves 
a logarithmic regret bound.
\\
\begin{algorithm}[tb]
   \caption{SC-RMSProp}
   \label{alg:sc_rmsprop}
\begin{algorithmic}
   \STATE {\bfseries Input:} $\theta_1 \in C$\,,\,$\delta_0=1$ , $v_0 = 0 \in \mathbb{R}^d$
   \FOR{$t=1$ {\bfseries to} T}
   \STATE $g_t \in \partial f_t(\theta_t)$
   \STATE $v_t = \beta_tv_{t-1}+(1-\beta_t)(g_t \odot g_t)$\
   \STATE Set $\epsilon_t = \frac{\delta_t}{t}$ where $\delta_{t,i}\leq \delta_{t-1,i}$ for $i \in [d]$ and $\alpha_t=\frac{\alpha}{t}$
   \STATE $A_t = \diag(v_t + \eps_t)$\;
   \STATE $\theta_{t+1} = P_C^{A_t}\big(\theta_t - \alpha_t A_t^{-1}g_t\big)$
   \ENDFOR
\end{algorithmic}
\end{algorithm}
Note that again there exist choices for the parameters of SC-RMSProp such that it reduces to SC-Adagrad.
The correspondence is given by the choice
\[ \beta_t = 1- \frac{1}{t}, \quad \alpha_t = \frac{\alpha}{t}, \quad\eps_t = \frac{\delta_t}{t},\]
for which again it follows $v_{t,i}=\frac{1}{t}\sum_{j=1}^t g^2_{j,i}$ with the same argument as for RMSProp. Please see Equation \eqref{eq:sc-adagrad} for the correspondence.  Moreover, with the same argument as for SC-Adagrad
we use a decay scheme for the damping factor
\[ \eps_{t,i} = \xi_2\frac{e^{-\xi_1 \,t\, v_{t,i}}}{t}, \quad i=1,\ldots,d.\, \text{ for } \xi_1 \geq 0\,,\, \xi_2 > 0\]

The analysis of SC-RMSProp is along the lines of SC-Adagrad with some overhead due to the structure of $v_t$.
\begin{lemma}\label{lem:dummy}
Let $\alpha_t=\frac{\alpha}{t}$, $1-\frac{1}{t}\leq \beta_t \leq 1- \frac{\gamma}{t}$ and $A_t$ as defined in SC-RMSProp, then it holds for all $T\geq 1$,
\begin{align*}
& \sum_{t=1}^{T}\frac{\alpha_t}{2}\inner{g_t, A_t^{-1}g_t}\leq \frac{\alpha}{2\gamma}\sum_{i=1}^d \log\Big(\frac{T\big(v_{T,i}+\eps_{T,i}\big)}{\epsilon_{1,i}}\Big)\\
&\quad + \frac{\alpha}{2\gamma}\sum_{t=2}^{T} \sum_{i=1}^d\frac{-t\epsilon_{t,i} +(t-1)\epsilon_{t-1,i} }{tv_{t,i}+t\epsilon_{t,i}}\\
& \quad +  \frac{\alpha}{2\gamma}\sum_{i=1}^d \frac{(1-\gamma)(1+\log T)}{\inf_{j \in [1,T]}jv_{j,i}+j\epsilon_{j,i}}
\end{align*}
\end{lemma}
\ifpaper
\begin{proof}
Using $\inner{x,Ax} = A \bullet \diag(xx^T)$ and with $g^2_{t,i} = \frac{v_{t,i}-\beta_t v_{t-1},i}{1-\beta_t}$ 
and using that $A_t$ is diagonal with $(A_t)_{ii}=v_{t,i}+\eps_{t,i}$, we get
\begin{align*}
& \sum_{t=1}^{T}\frac{\alpha}{2} ((tA_{t})^{-1} \bullet  \diag(g_tg_t^T))\\
&=\frac{\alpha}{2(1 - \beta_{1})}((A_{1})^{-1} \bullet  (A_1 - \diag(\epsilon_1)))\\
& + \sum_{t=2}^{T}\frac{\alpha}{2}  \Big((tA_{t})^{-1}\bullet  \Big(\frac{A_{t} - \diag(\epsilon_t) }{(1-\beta_{t})}\Big)\Big)\\
& - \sum_{t=2}^{T}\frac{\alpha}{2}  \Big((tA_{t})^{-1}\bullet  \Big(\frac{\beta_{t}A_{t-1} - \beta_{t}\diag(\epsilon_{t-1})}{(1-\beta_{t})}\Big)\Big)\\ 
&\leq \frac{\alpha}{2\gamma}((A_{1})^{-1} \bullet  (A_1 - \diag(\epsilon_1)))\\ & +\sum_{t=2}^{T}\frac{\alpha}{2} \Big((tA_{t})^{-1} \bullet  \Big(\frac{tA_{t}-t\beta_{t}A_{t-1}}{\gamma}\Big)\Big)\\
& + \sum_{t=2}^{T}\frac{\alpha}{2} \Big((tA_{t})^{-1} \bullet  \Big(\frac{- \diag(\epsilon_t) + \beta_{t}\diag(\epsilon_{t-1})}{(1-\beta_{t})}\Big)\Big)\\
&\leq \frac{\alpha}{2\gamma}\log\Big(\frac{|A_1|}{|\diag(\epsilon_1)|}\Big) +\frac{\alpha}{2\gamma}\sum_{t = 2}^T \log\Big(\frac{|tA_t|}{|t\beta_tA_{t-1}|} \Big)\\
& +  \sum_{t=2}^{T}\frac{\alpha}{2} \Big((tA_{t})^{-1} \bullet  \Big(\frac{- \diag(\epsilon_t) + \beta_{t}\diag(\epsilon_{t-1})}{(1-\beta_{t})}\Big)\Big)\\
&\leq \frac{\alpha}{2\gamma}\log\Big(\frac{|A_1|}{|\diag(\epsilon_1)|}\Big) +\frac{\alpha}{2\gamma}\sum_{t = 2}^T \log\Big(\frac{|tA_t|}{|(t-1)A_{t-1}|} \Big) \\
& +  \sum_{t=2}^{T}\frac{\alpha}{2} \Big((tA_{t})^{-1} \bullet  \Big(\frac{- \diag(\epsilon_t) + \beta_{t}\diag(\epsilon_{t-1})}{(1-\beta_{t})}\Big)\Big)
\end{align*}
In the first inequality we use $\frac{1}{1-\beta_t} \leq \frac{t}{\gamma}$. In the last step we use, that $\forall t > 1$, $\frac{1}{t\beta_t} \leq  \frac{1}{t-1}$. Finally, by upper bounding
the last term with Lemma \ref{lem:what}
\begin{align*}
& \sum_{t=1}^{T}\frac{\alpha_t}{2}\inner{g_t, A_t^{-1}g_t}\leq \frac{\alpha}{2\gamma}\log\Big(\frac{|TA_T|}{|\diag(\epsilon_1)|}\Big)\\
&\quad + \frac{\alpha}{2\gamma}\sum_{t=2}^{T} \sum_{i=1}^d\frac{-t\epsilon_{t,i} +(t-1)\epsilon_{t-1,i} }{tv_{t,i}+t\epsilon_{t,i}}\\
& \quad +  \frac{\alpha}{2\gamma}\sum_{i=1}^d \frac{(1-\gamma)(1+\log T)}{\inf_{j \in [1,T]}jv_{j,i}+j\epsilon_{j,i}}
\end{align*}
We note that
\begin{align*}
\log(|T A_T|) &= \log\Big( \prod_{i=1}^d (T (v_{T,i}+\epsilon_{T,i} ))\Big)\\
&= \sum_{i=1}^d \log(T (v_{T,i}+\epsilon_{T,i}) ),
\end{align*}
and similar, $\log(|\diag(\eps_1)|)=\sum_{i=1}^d \log(\eps_{1,i})$.
\end{proof}
\fi
Note that for $\gamma=1$ and the choice $\eps_t=\frac{\delta_t}{t}$ this reduces to the result of Lemma \ref{lem:sc-lem}.
\begin{lemma}\label{lem:what}
Let $\eps_t \leq \frac{1}{t}$  and $1-\frac{1}{t}\leq \beta_t \leq 1-\frac{\gamma}{t}$ for some $1\geq \gamma>0$. Then it holds,
\begin{align*}
& \sum_{t=2}^{T}\frac{\alpha}{2} \Big((tA_{t})^{-1} \bullet  \Big(\frac{- \diag(\epsilon_t) + \beta_{t}\diag(\epsilon_{t-1})}{(1-\beta_{t})}\Big)\Big)\\
& \leq  \frac{\alpha}{2\gamma}\sum_{t=2}^{T} \sum_{i=1}^d\frac{-t\epsilon_{t,i} +(t-1)\epsilon_{t-1,i} }{tv_{t,i}+t\epsilon_{t,i}}\\
& \quad +  \frac{\alpha}{2\gamma}\sum_{i=1}^d \frac{(1-\gamma)(1+\log T)}{\inf_{j \in [1,T]}jv_{j,i}+j\epsilon_{j,i}}
\end{align*}
\end{lemma}

\ifpaper
\begin{proof} Using $\frac{\gamma}{t}\leq 1-\beta_t \leq \frac{1}{t}$, we get
\begin{align*}
& \sum_{t=2}^{T}\frac{\alpha}{2} \Big((tA_{t})^{-1} \bullet  \Big(\frac{- \diag(\epsilon_t) + \beta_{t}\diag(\epsilon_{t-1})}{(1-\beta_{t})}\Big)\Big)\\
&\leq \frac{\alpha}{2}\sum_{t=2}^{T}  \Big((tA_{t})^{-1} \bullet  \Big(\frac{- t\diag(\epsilon_t) + t\beta_{t}\diag(\epsilon_{t-1})}{\gamma}\Big)\Big)\\
&= \frac{\alpha}{2\gamma}\sum_{t=2}^{T} \sum_{i=1}^d\frac{-t\epsilon_{t,i}+t\beta_t\epsilon_{t-1,i}}{tv_{t,i}+t\epsilon_{t,i}}\\
&= \frac{\alpha}{2\gamma}\sum_{t=2}^{T} \sum_{i=1}^d\frac{-t\epsilon_{t,i} +(t-1)\epsilon_{t-1,i} + (t\beta_{t}-(t-1))\epsilon_{t-1,i}}{tv_{t,i}+t\epsilon_{t,i}}\\
& \leq  \frac{\alpha}{2\gamma}\sum_{t=2}^{T} \sum_{i=1}^d\frac{-t\epsilon_{t,i} +(t-1)\epsilon_{t-1,i} }{tv_{t,i}+t\epsilon_{t,i}}\\
&\quad +  \frac{\alpha}{2\gamma}\sum_{t=2}^{T} \sum_{i=1}^d\frac{(1-\gamma)\epsilon_{t-1,i}}{tv_{t,i}+t\epsilon_{t,i}}\\
& \leq   \frac{\alpha}{2\gamma}\sum_{t=2}^{T} \sum_{i=1}^d\frac{-t\epsilon_{t,i} +(t-1)\epsilon_{t-1,i} }{tv_{t,i}+t\epsilon_{t,i}}\\
 &\quad+  \frac{\alpha}{2\gamma}\sum_{i=1}^d \frac{1-\gamma}{\inf_{j \in [2,T]}jv_{j,i}+j\epsilon_{j,i}} \sum_{t=2}^T\frac{1}{t}\\
& \leq  \frac{\alpha}{2\gamma}\sum_{t=2}^{T} \sum_{i=1}^d\frac{-t\epsilon_{t,i} +(t-1)\epsilon_{t-1,i} }{tv_{t,i}+t\epsilon_{t,i}}\\
&\quad +  \frac{\alpha}{2\gamma}\sum_{i=1}^d \frac{1-\gamma}{\inf_{j \in [2,T]}jv_{j,i}+j\epsilon_{j,i}} (1+\log T )
\end{align*}
\end{proof}
\fi

\begin{theorem}
Let Assumptions \textbf{A1}, \textbf{A2} hold and let ${\theta_t}$ be the sequence generated by SC-RMSProp in Algorithm \ref{alg:sc_rmsprop}, where $g_t \in \partial f_t(\theta_t)$  and $f_t:C \rightarrow \R$ is an arbitrary $\mu$-strongly convex function $(\mu \in \R^d_+)$ with $\alpha_t=\frac{\alpha}{t}$ for some $\alpha\geq \frac{(2-\gamma)G_\infty^2}{2 \min_i \mu_i}$ and $1-\frac{1}{t}\, \leq \, \beta_{t} \, \leq 1-\frac{\gamma}{t}$  for some $0<\gamma\leq 1$. Furthermore, set $\eps_t = \frac{\delta_t}{t}$ and assume $1\geq \delta_{t,i} > 0$  and $\delta_{t,i} \leq \delta_{t-1,i} \forall t \in [T], \forall i \in [d]$, then
the regret of SC-RMSProp can be upper bounded for $T\geq 1$ as
\begin{align*}
& R(T)\, \leq \,\frac{D_{\infty}^2\tr(\diag(\delta_1))}{2\alpha}+    \frac{\alpha}{2\gamma}\sum_{i=1}^d \log\Big(\frac{T v_{T,i}+\delta_{T,i}}{\delta_{1,i}}\Big)\\
& + \frac{1}{2}\sum_{i=1}^{d} \inf_{t \in [T]}\Big(\frac{(\theta_{t,i}-\theta_{i}^*)^2}{\alpha}- \frac{\alpha}{\gamma (tv_{t,i}+t\epsilon_{t,i})}\Big) (\delta_{T,i}-\delta_{1,i}) \\
& +  \frac{\alpha}{2\gamma}\sum_{i=1}^d \frac{(1-\gamma)(1+\log T)}{\inf_{j \in [1,T]}jv_{j,i}+j\epsilon_{j,i}}
\end{align*}
\end{theorem}
\ifpaper
\begin{proof}
We rewrite the regret bound with the definition of $\mathbf{\mu}$-strongly convex functions as 
\begin{align*}
& R(T)=\sum_{t=1}^T (f_t(\theta_t)-f_t(\theta^*))\\
& \leq \sum_{t=1}^T \inner{g_t,\theta_{t}-\theta^*} - \sum_{t=1}^T \norm{\theta_t-\theta^*}^2_{\diag(\mu)}
\end{align*}
Using the non-expansiveness of the weighted projection, we get
\begin{align*}
& \norm{\theta_{t+1}-\theta^*}^2_{A_t}\\
 &= \norm{P_C^{A_t}\big(\theta_t-\alpha_t A_t^{-1} g_t\big)-\theta^*}^2_{A_t}\\
&\leq  \norm{\theta_t-\alpha_t A_t^{-1} g_t-\theta^*}^2_{A_t}\\
&\leq \|\theta_t - \theta^*\|^2_{A_t} - 2\alpha_t\inner{g_t, \theta_t - \theta^*}+ \alpha_t^2\inner{g_t,A_t^{-1}g_t}
\end{align*}
This yields
\begin{align*}
& \inner{g_t,\theta_t - \theta^*}\\
&\leq \frac{\|\theta_t-\theta^*\|^2_{A_t} - \|\theta_{t+1}-\theta^*\|^2_{A_t}}{2\alpha_t} + \frac{\alpha_t}{2}\inner{g_t,A_t^{-1}g_t}
\end{align*}
Hence we can upper bound the regret as follows
\begin{align*}
& R(T)\\
&\leq \sum_{t=1}^{T}\frac{\|\theta_t-\theta^*\|^2_{A_t} - \|\theta_{t+1}-\theta^*\|^2_{A_t}}{2\alpha_t} \\
& + \sum_{t=1}^{T}\frac{\alpha_t}{2}\inner{g_t,A_t^{-1}g_t} - \sum_{t=1}^T \norm{\theta_t-\theta^*}^2_{\diag(\mu)}\\
&\leq\frac{\|\theta_1 - \theta^*\|^2_{A_1}}{2\alpha_1}+\sum_{t=2}^{T}\Big(\frac{\|\theta_t-\theta^*\|^2_{A_t}}{2\alpha_t} - \frac{\|\theta_t-\theta^*\|^2_{A_{t-1}}}{2\alpha_{t-1}}\Big)  \\
& - \frac{\|\theta_{T+1}-\theta^*\|^2_{A_{T}}}{2\alpha_t}+ \sum_{t=1}^{T}\frac{\alpha_t}{2}\inner{g_t,A_t^{-1}g_t}\\
& - \sum_{t=1}^T \norm{\theta_t-\theta^*}^2_{\diag(\mu)}\\
&\leq \frac{\|\theta_1 - \theta^*\|^2_{A_1}}{2\alpha_1}+\sum_{t=2}^{T}\frac{\|\theta_t-\theta^*\|^2_{tA_t - (t-1)A_{t-1} } }{2\alpha} \\
& + \sum_{t=1}^{T}\frac{\alpha_t}{2}\inner{g_t, A_t^{-1}g_t} - \sum_{t=1}^T \norm{\theta_t-\theta^*}^2_{\diag(\mu)}
\end{align*}
Now on imposing the following condition
\begin{equation}\label{eq:one}
\begin{split}
tA_t - (t-1)A_{t-1} - 2\alpha\, \diag(\mu)  \\
\preceq \diag(\delta_t) - \diag(\delta_{t-1})\, \forall t \geq 2
\end{split}
\end{equation}
\begin{equation}\label{eq:two}
A_1 - 2\alpha \diag(\mu) \preceq \diag(\delta_1)
\end{equation}
Note that with $\epsilon_{t,i}=\frac{\delta_{t,i}}{t}$ and $\delta_{t,i}\leq \delta_{t-1,i}$ for all $t\geq 1$ and $i \in [d]$, it holds $t\epsilon_{t,i} \leq (t-1)\epsilon_{t-1,i}$. We show regarding the  inequality in \eqref{eq:one}
\begin{align*}
& tv_{t,i} + t\epsilon_{t,i} - (t-1)v_{t-1,i} -(t-1)\epsilon_{t-1,i} -2\alpha\mu_i \\
& = t\beta_{t,i}v_{t-1,i} + t\epsilon_{t,i} + t(1-\beta_t)g_{t,i}^2 -(t-1)\epsilon_{t-1,i} \\
&- (t-1)v_{t-1,i} -2\alpha\mu_i \\
& =(t\beta_{t,i} - (t-1))v_{t-1,i} + t(1-\beta_t)g_{t,i}^2 -2\alpha\mu_i \\
&+ t\epsilon_{t,i} - (t-1)\epsilon_{t-1,i}\\
& \leq (1-\gamma)v_{t-1,i} + g_{t,i}^2 - 2\alpha \mu_i + t\epsilon_{t,i} - (t-1)\epsilon_{t-1,i}\\
& \leq (1-\gamma)G_{\infty}^2 + G_{\infty}^2 - 2\alpha \mu_i + t\epsilon_{t,i} - (t-1)\epsilon_{t-1,i}\\
& = (2-\gamma) G_{\infty}^2 - 2\alpha \mu_i + t\epsilon_{t,i} - (t-1)\epsilon_{t-1,i}\\
& \leq t\epsilon_{t,i} - (t-1)\epsilon_{t-1,i},
\end{align*}
where the last inequality follows by  choosing $\alpha \geq \frac{G_{\infty}^2}{2\underset{i}{min}\mu_i}(2-\gamma)$ and we have used that
\begin{align*}
 v_{t,i} &= \sum_{j=1}^t  (1-\beta_j) \prod_{k=j+1}^t \beta_k g^2_{j,i}\\
           &\leq G_{\infty}^2  \sum_{j=1}^t \Big( \prod_{k=j+1}^t \beta_k -  \prod_{k=j}^t \beta_k \Big)\\
           &\leq G_{\infty}^2 \Big( 1 - \prod_{k=1}^t \beta_k\Big)\leq G_\infty^2
\end{align*}
The second inequality \eqref{eq:two} holds easily with the given choice of $\alpha$.
Choosing some $\beta_t = 1 - \frac{\gamma}{t}$
\begin{align*}
& tv_{t,i}  = t\beta_{t,i}v_{t-1,i} + t(1-\beta_t)g_{t,i}^2\\
& \geq (t-1)v_{t-1,i} + t(1-\beta_t)g_{t,i}^2 \geq (t-1)v_{t-1,i}
\end{align*}
Hence $\delta_{t,i} \leq \delta_{t-1,i}$ where $\delta_{t,i} = e^{-\epsilon\, t\,v_{t,i}}$ for $\epsilon > 0$. With $\epsilon_{t,i} = \frac{\delta_{t,i}}{t}$ we have  $t\epsilon_t \leq (t-1)\epsilon_{t-1}$.
\begin{align*}
&R(T)\\
& \leq \frac{\|\theta_1 - \theta^*\|^2_{\diag(\delta_1)}}{2\alpha}+ \sum_{t=2}^{T}\frac{\|\theta_t - \theta^*\|_{\diag(\delta_t)-\diag(\delta_{t-1})}}{2\alpha}\\
&\quad + \sum_{t=1}^{T}\frac{\alpha_t}{2}\inner{g_t, A_t^{-1}g_t}\\
& = \frac{\|\theta_1 - \theta^*\|^2_{\diag(\delta_1)}}{2\alpha}+ \sum_{t=2}^{T}\sum_{i=1}^{d}\frac{(\theta_{t,i}-\theta_{i}^*)^2(\delta_{t,i}-\delta_{t-1,i})}{2\alpha}\\
&\quad +\sum_{t=1}^{T} \frac{\alpha_t}{2}\inner{g_t, A_t^{-1}g_t}\\
&\leq\frac{D_{\infty}^2\tr(\diag(\delta_1))}{2\alpha}+ \sum_{t=1}^{T}\frac{\alpha_t}{2}\inner{g_t, A_t^{-1}g_t}\\
& \quad + \sum_{t=2}^{T}\sum_{i=1}^{d}\frac{(\theta_{t,i}-\theta_{i}^*)^2(\delta_{t,i}-\delta_{t-1,i})}{2\alpha}\\
&\leq \frac{D_{\infty}^2\tr(\diag(\delta_1))}{2\alpha}+    \frac{\alpha}{2\gamma}\sum_{i=1}^d \log\Big(\frac{T v_{T,i}+\delta_{T,i}}{\delta_{1,i}}\Big)\\
& \quad + \frac{1}{2}\sum_{t=2}^{T}\sum_{i=1}^{d}\Bigg(\frac{(\theta_{t,i}-\theta_{i}^*)^2(\delta_{t,i}-\delta_{t-1,i})}{\alpha}- \frac{\alpha(\delta_{t,i}-\delta_{t-1,i})}{\gamma (tv_{t,i}+t\epsilon_{t,i})}\Bigg)\\
& \quad +  \frac{\alpha}{2\gamma}\sum_{i=1}^d \frac{(1-\gamma)(1+\log T)}{\inf_{j \in [1,T]}jv_{j,i}+j\epsilon_{j,i}}\\
&\leq \frac{D_{\infty}^2\tr(\diag(\delta_1))}{2\alpha}+    \frac{\alpha}{2\gamma}\sum_{i=1}^d \log\Big(\frac{T v_{T,i}+\delta_{T,i}}{\delta_{1,i}}\Big)\\
& \quad + \frac{1}{2}\sum_{i=1}^{d} \inf_{t \in [T]}\Big(\frac{(\theta_{t,i}-\theta_{i}^*)^2}{\alpha}- \frac{\alpha}{\gamma (tv_{t,i}+t\epsilon_{t,i})}\Big) (\delta_{T,i}-\delta_{1,i}) \\
& \quad +  \frac{\alpha}{2\gamma}\sum_{i=1}^d \frac{(1-\gamma)(1+\log T)}{\inf_{j \in [1,T]}jv_{j,i}+j\epsilon_{j,i}}
\end{align*}
where we have used Lemma \ref{lem:dummy} in the last inequality and $\eps_{t,i}=\frac{\delta_{t,i}}{t}$ for $i \in [d]$ and $t\geq 1$.
\end{proof}
\fi
\begin{figure*}[ht]
\centering     
\subfigure[CIFAR10]{\label{fig:mlp_cifar10}\includegraphics[width=0.32\textwidth]{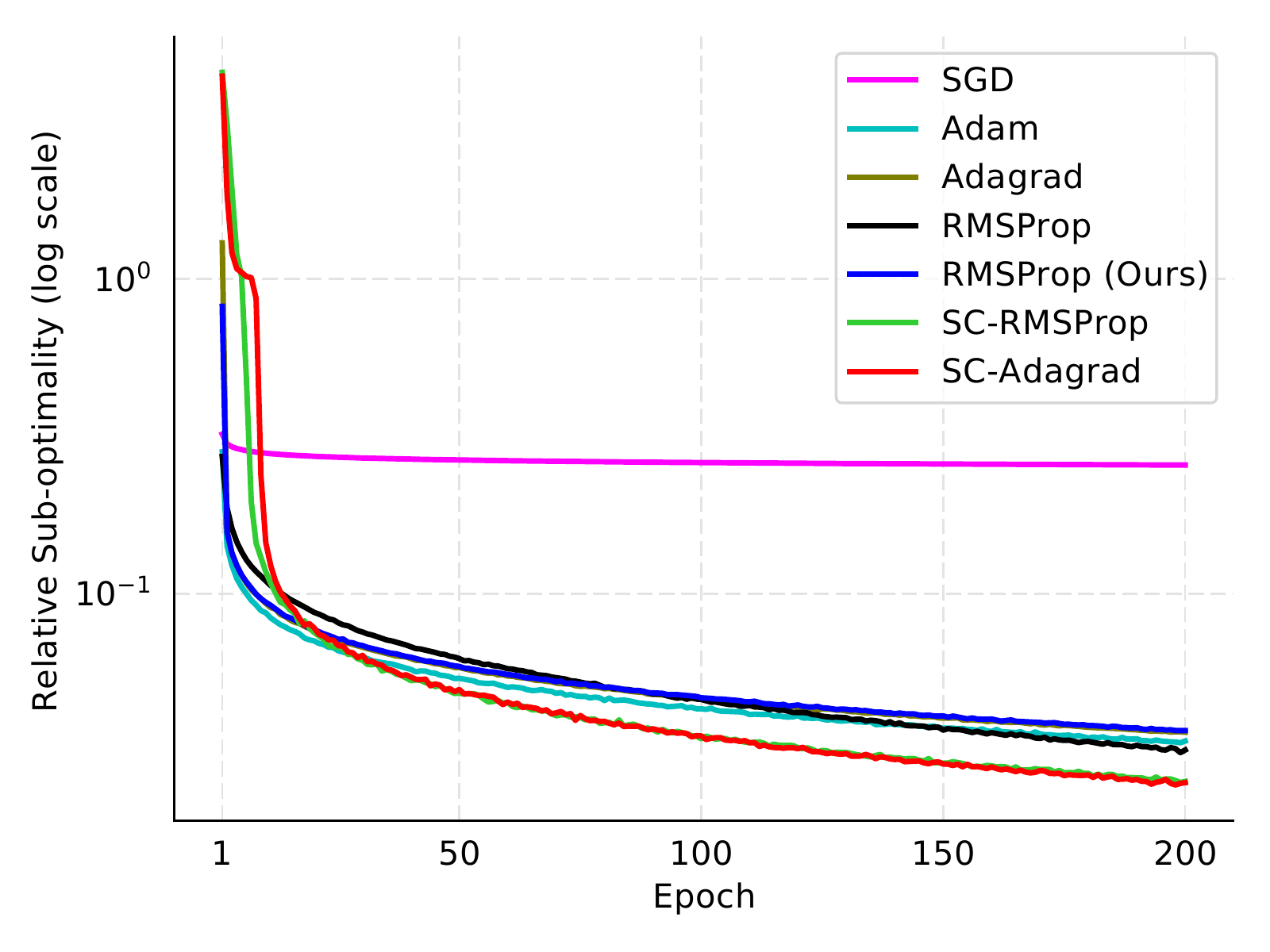}}
\subfigure[CIFAR100]{\label{fig:mlp_cifar100}\includegraphics[width=0.32\textwidth]{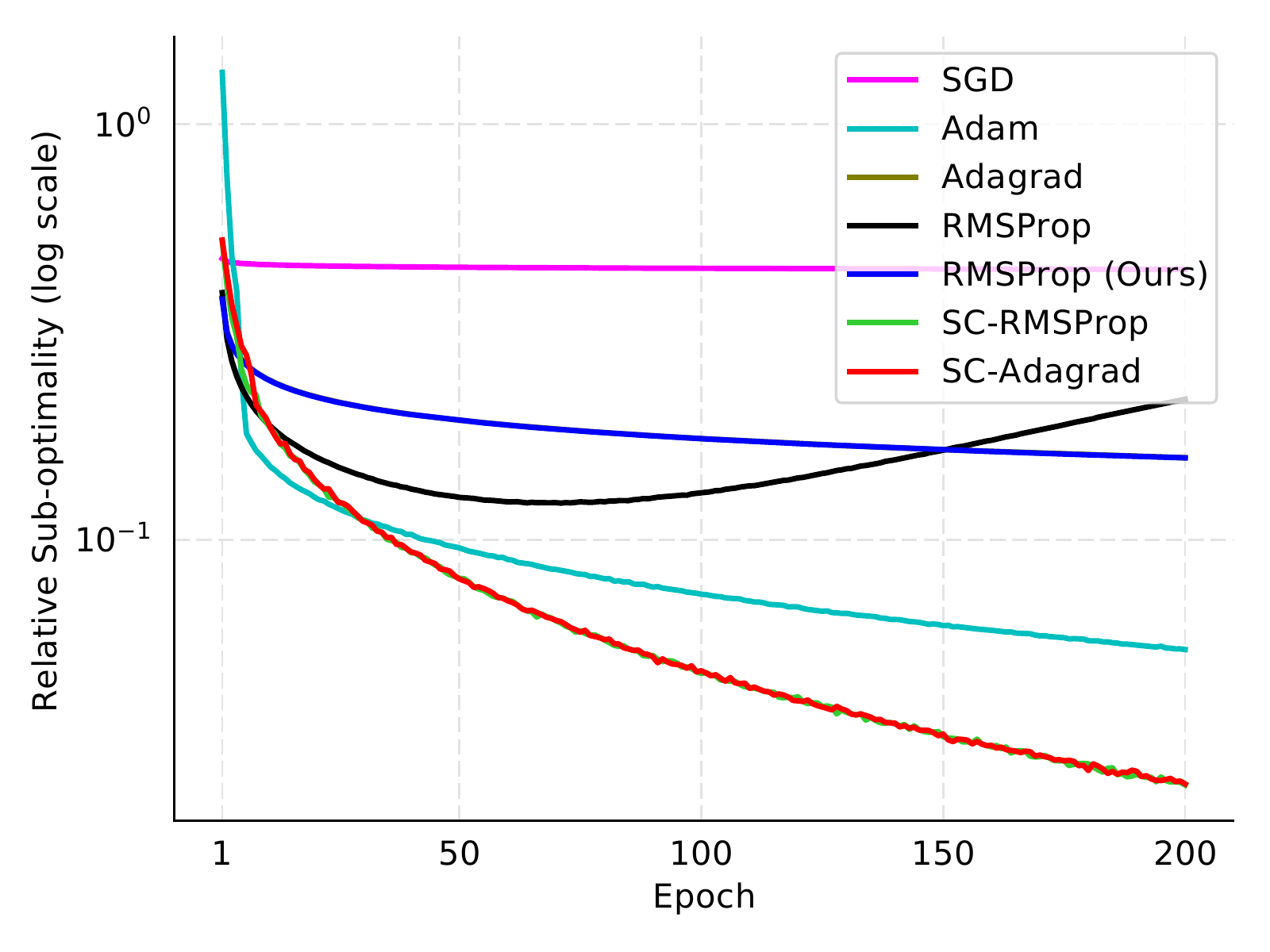}}
\subfigure[MNIST]{\label{fig:mlp_mnist}\includegraphics[width=0.32\textwidth]{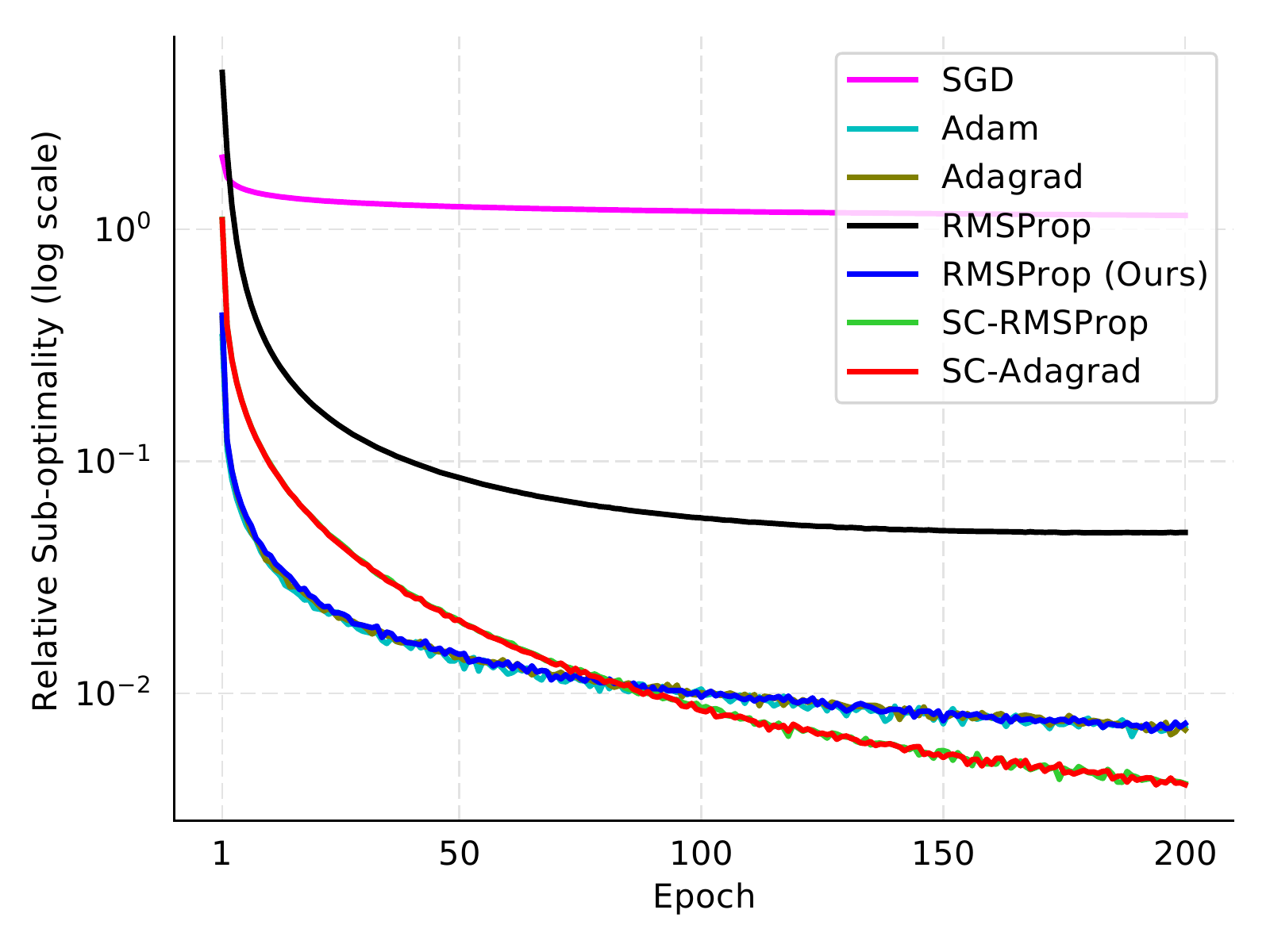}}
\caption{Relative Suboptimality  vs Number of Epoch for L2-Regularized Softmax Regression}
\label{fig:logistic_test}
\end{figure*}
\begin{figure*}[ht]
\centering     
\subfigure[CIFAR10]{\label{fig:test_online_logistic_cifar10}\includegraphics[width=0.32\textwidth]{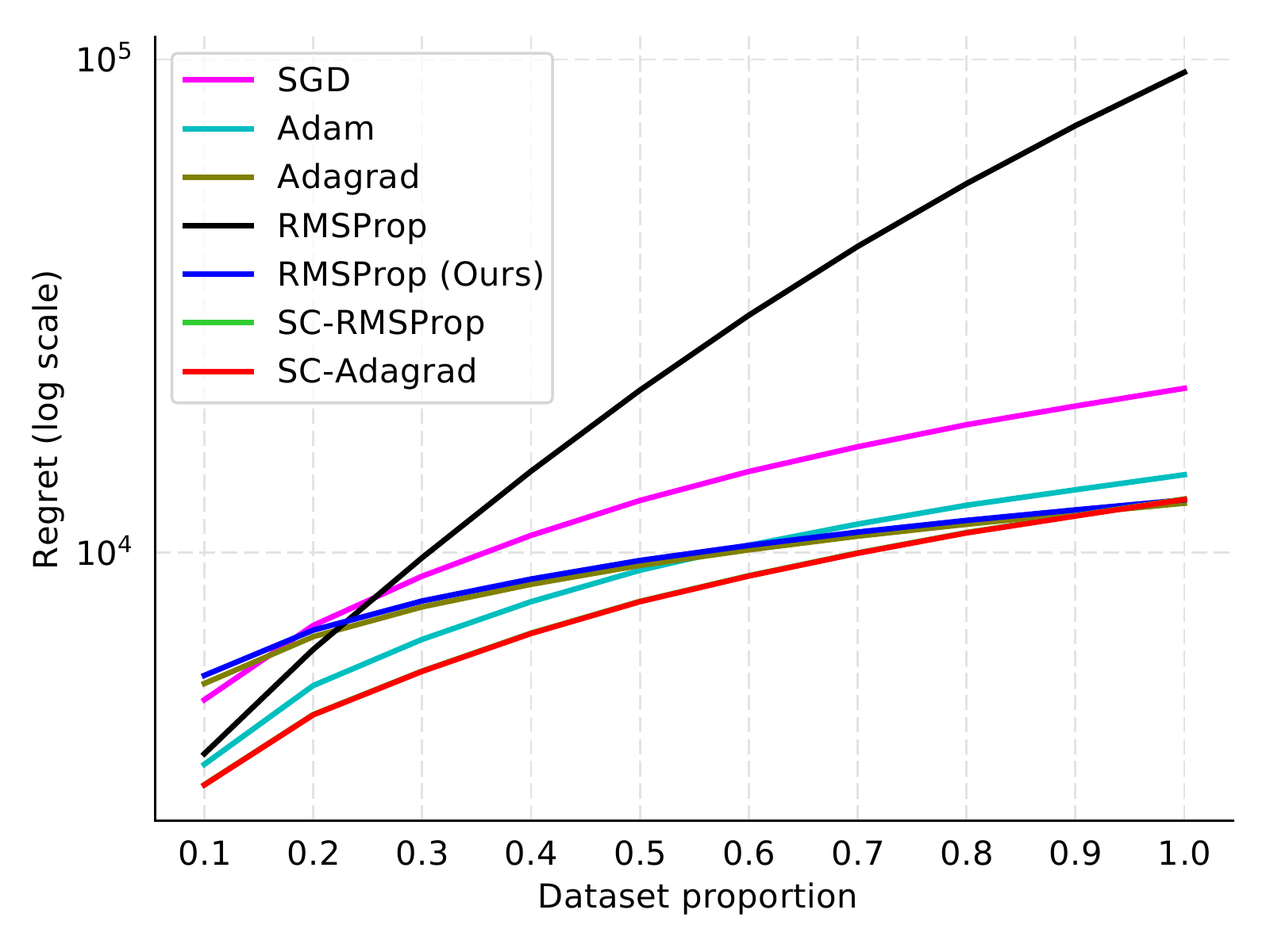}}
\subfigure[CIFAR100]{\label{fig:test_online_logistic_cifar100}\includegraphics[width=0.32\textwidth]{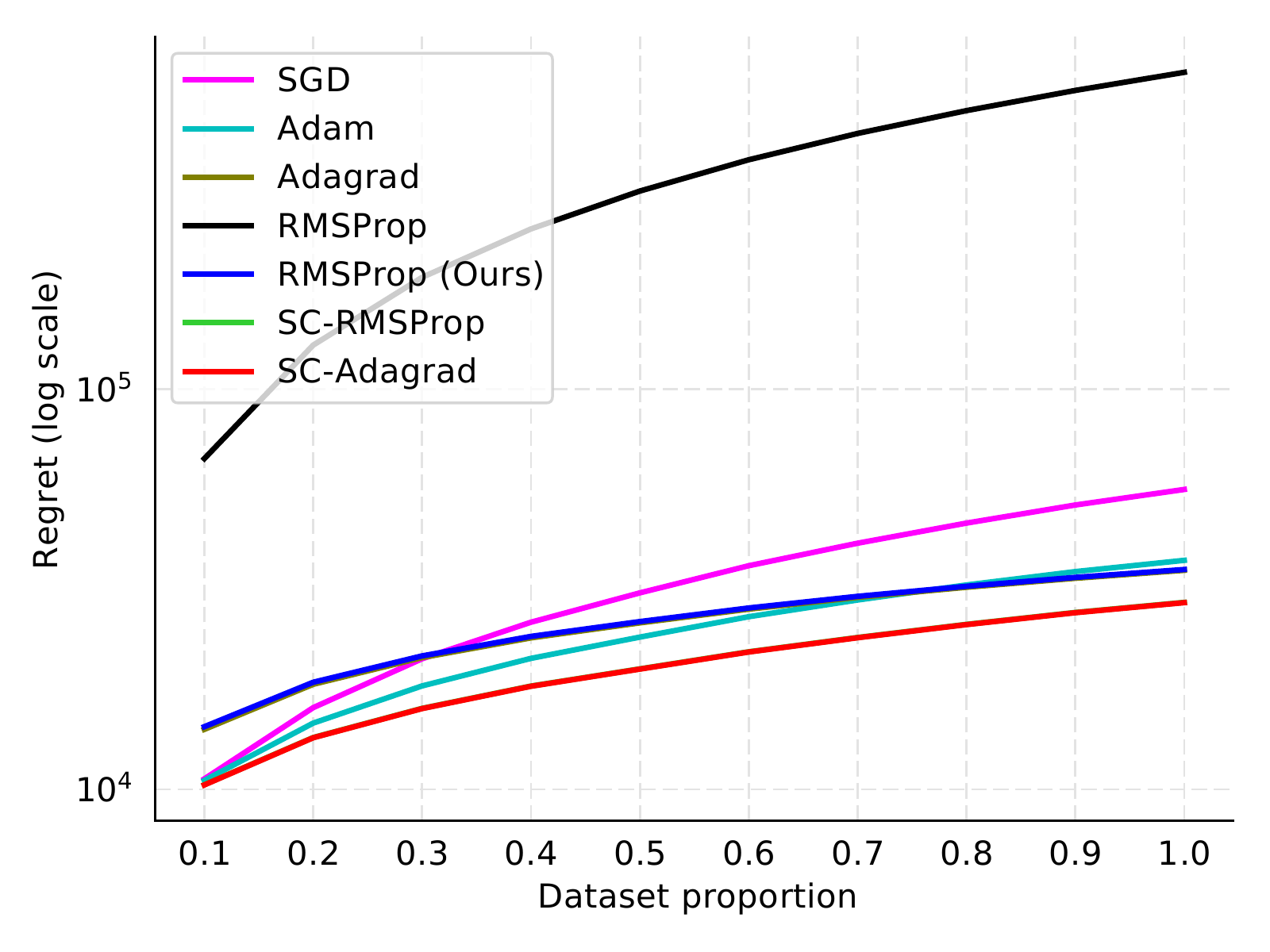}}
\subfigure[MNIST]{\label{fig:test_online_logistic_mnist}\includegraphics[width=0.32\textwidth]{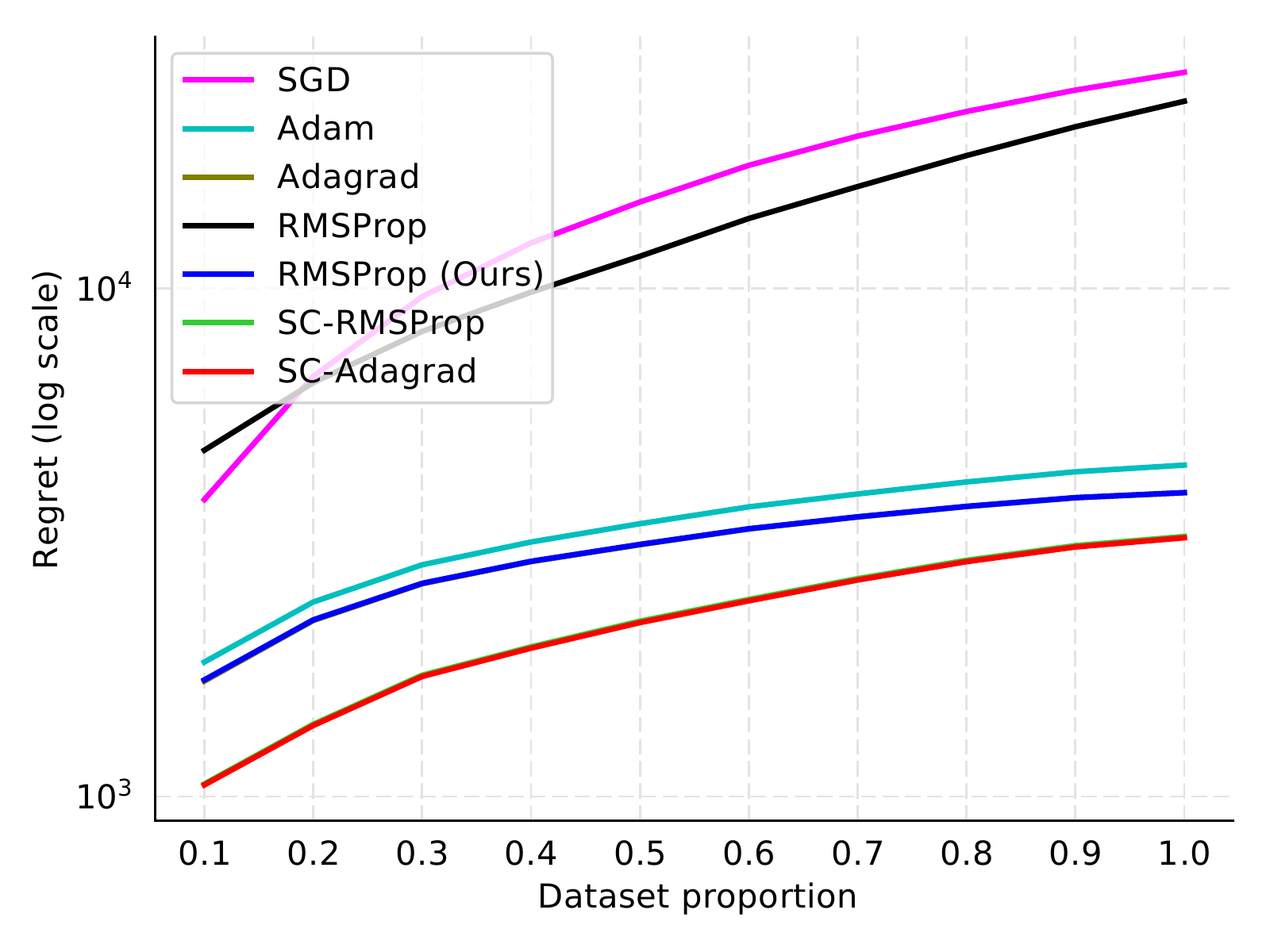}}
\caption{Regret (log scale) vs Dataset Proportion for Online L2-Regularized Softmax Regression }
\label{fig:test_online_logistic_regret}
\end{figure*}
Note that the regret bound reduces for $\gamma=1$ to that of SC-Adagrad. For $0<\gamma<1$ a comparison between
the bounds is not straightforward as the $v_{t,i}$ terms cannot be compared. It is an interesting future research question
whether it is possible to show that one scheme is better than the other one potentially dependent on the problem characteristics.

\begin{figure*}[ht]
\centering     
\subfigure[CIFAR10]{\label{fig:val_cnn_cifar10_test_acc}\includegraphics[width=0.32\textwidth]{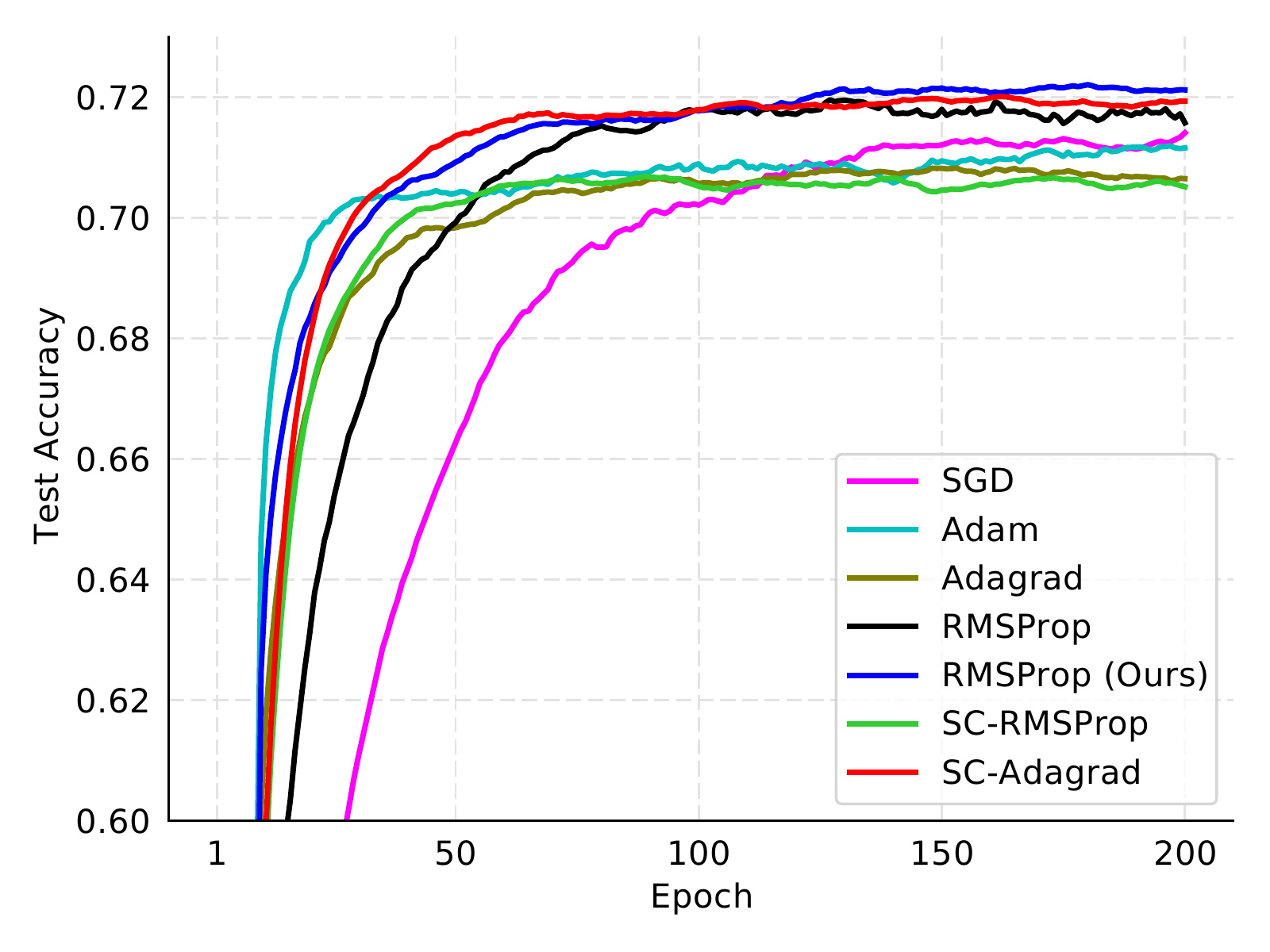}}
\subfigure[CIFAR100]{\label{fig:val_cnn_cifar100_test_acc}\includegraphics[width=0.32\textwidth]{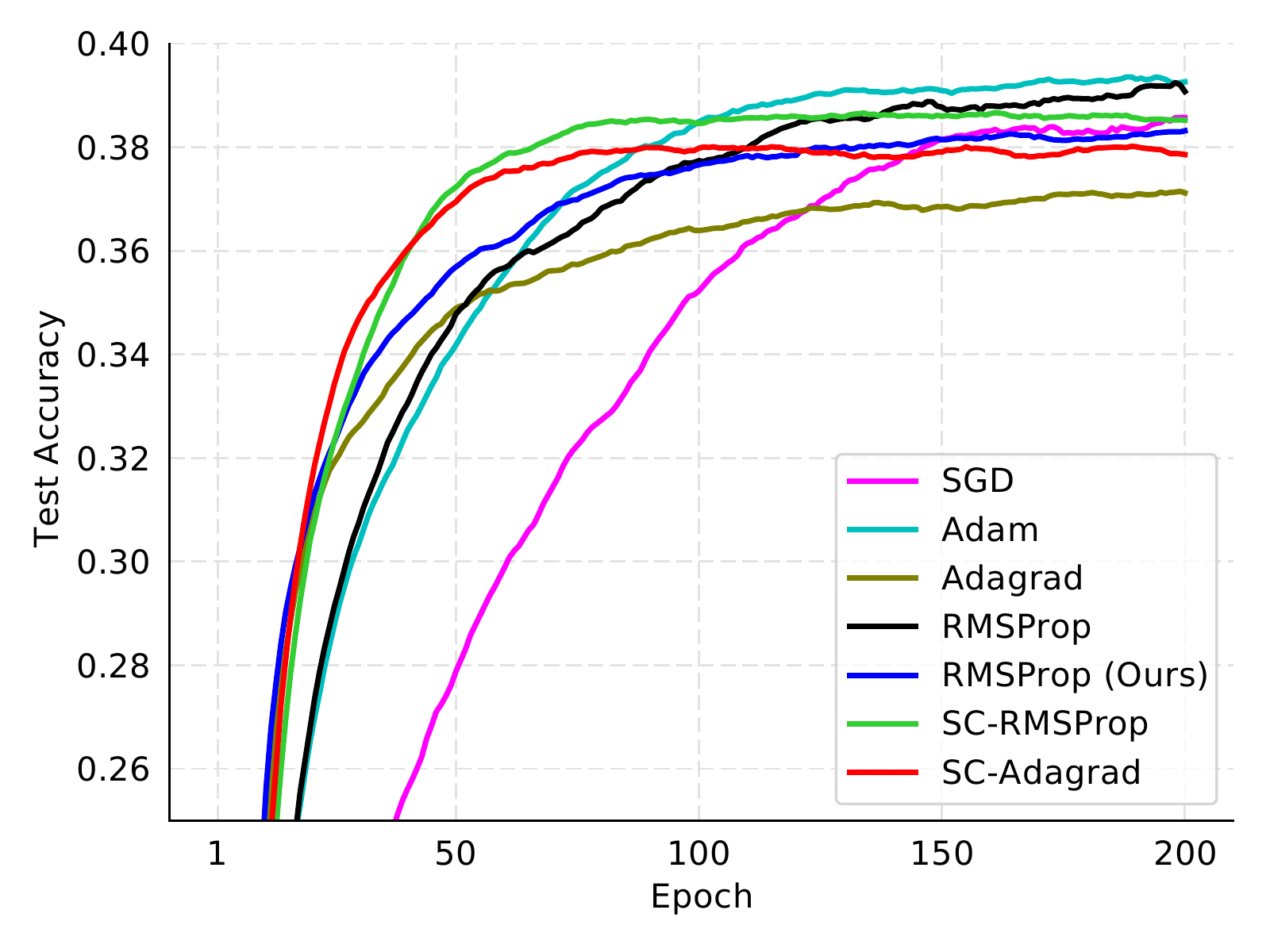}}
\subfigure[MNIST]{\label{fig:val_cnn_mnist_test_acc}\includegraphics[width=0.32\textwidth]{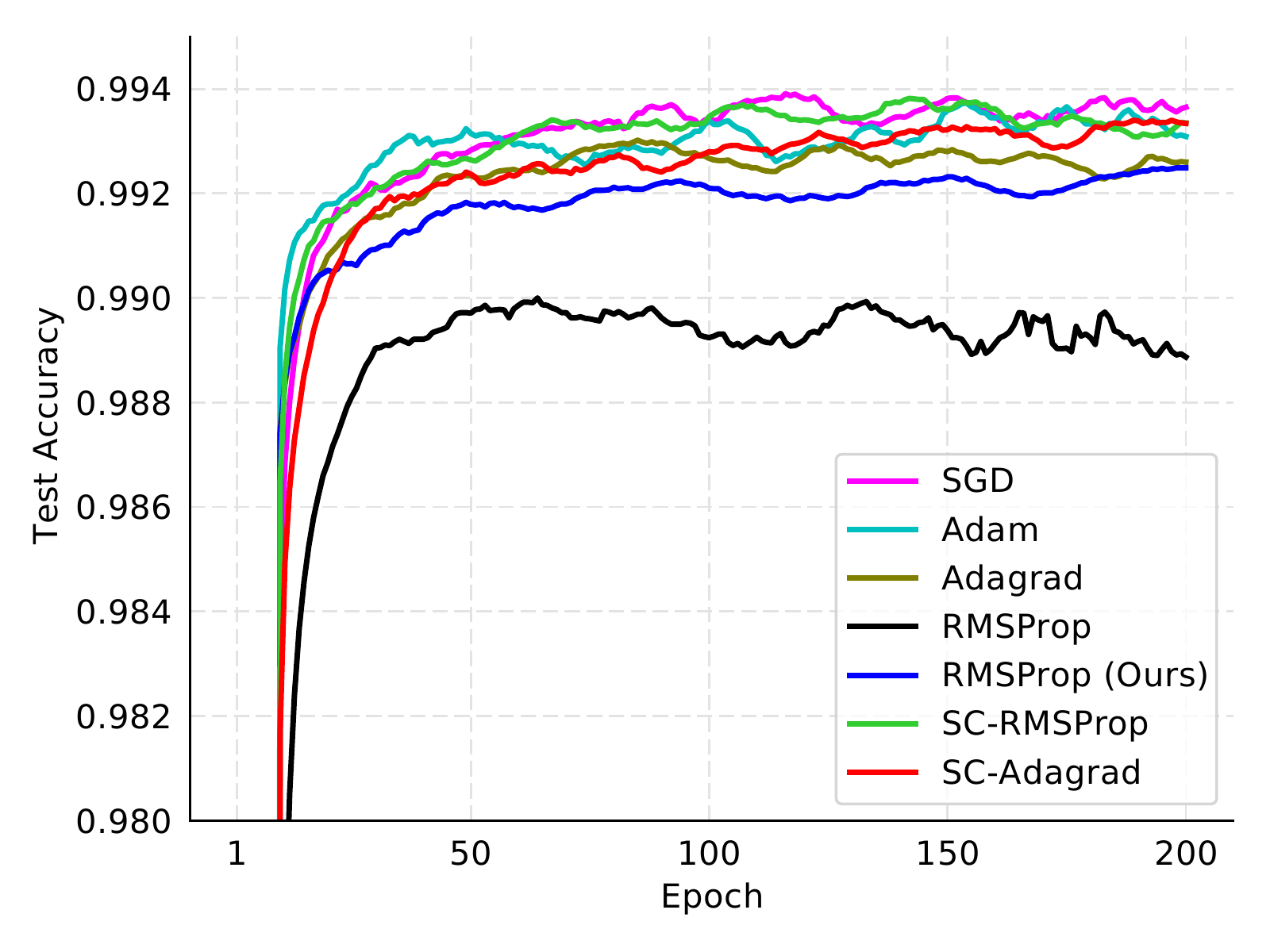}}
\caption{Test Accuracy vs Number of Epochs for 4-layer CNN}
\label{fig:val_cnn_test_acc}
\end{figure*}

\begin{figure*}[ht]

\centering     
\subfigure[Training Objective]{\label{fig:resnet_cifar10_loss}\includegraphics[width=0.35\textwidth]{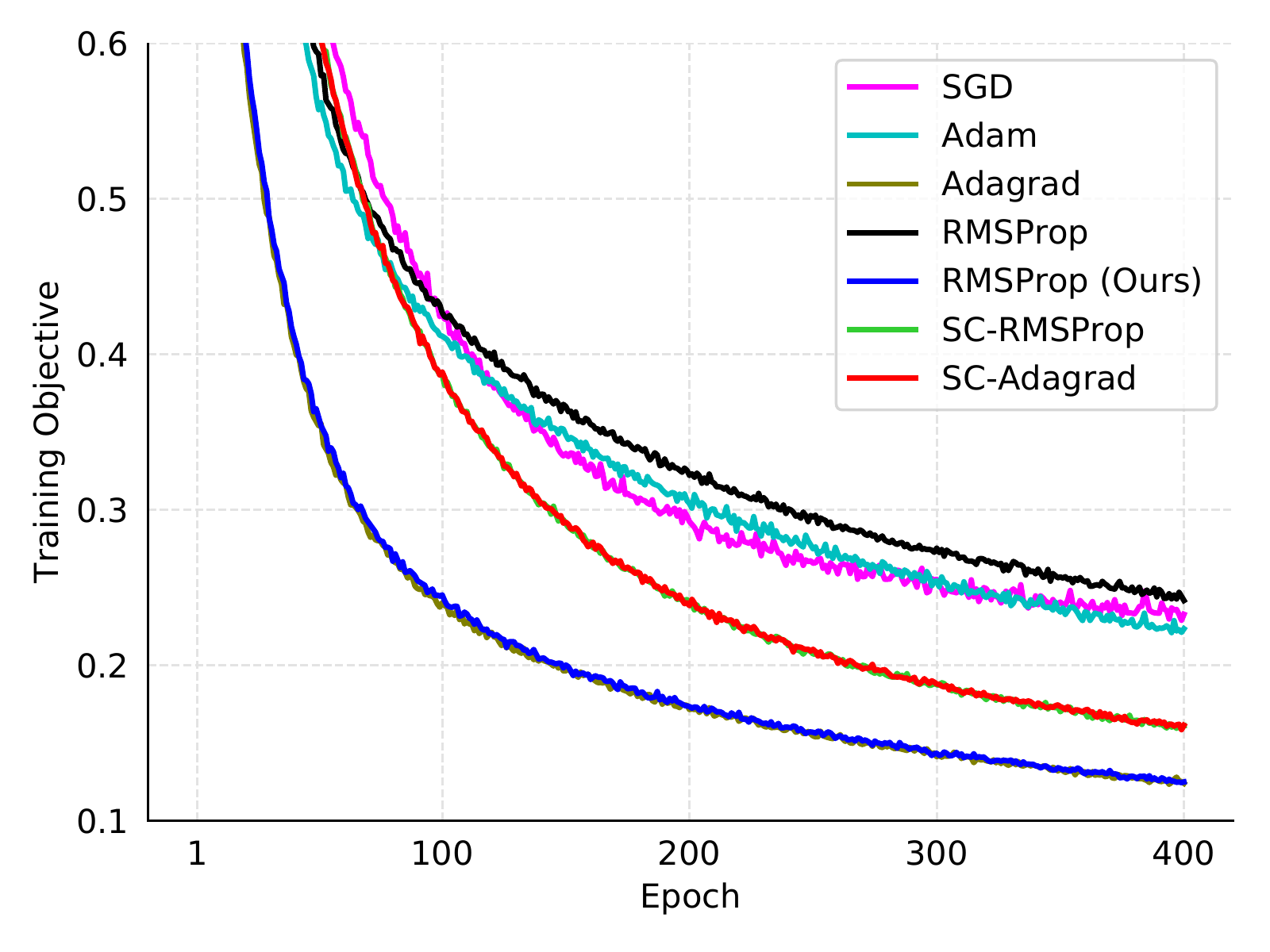}}
\subfigure[Test Accuracy]{\label{fig:resnet_cifar10_val_acc_mvg_avg}\includegraphics[width=0.35\textwidth]{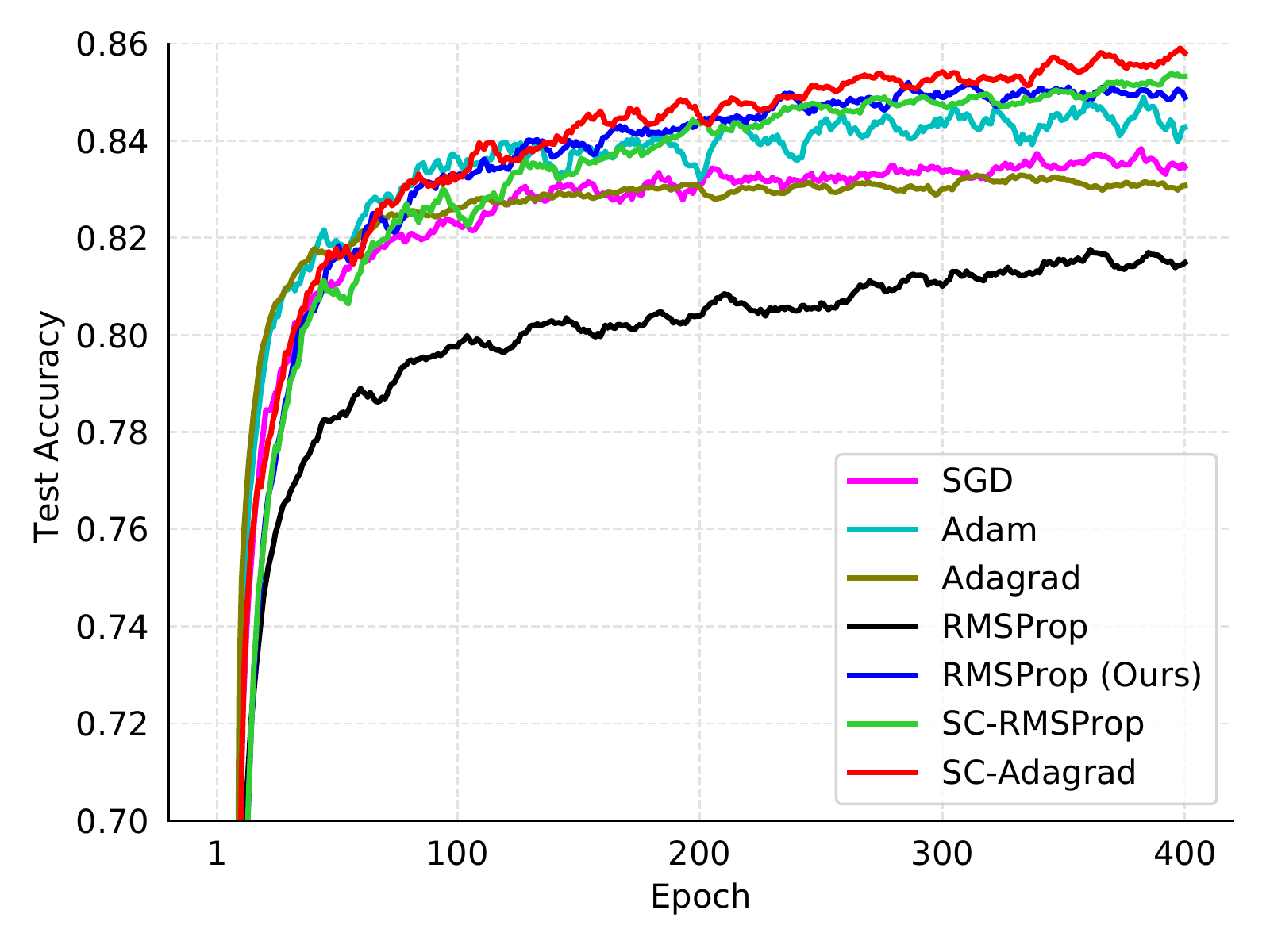}}
\caption{Plots of ResNet-18 on CIFAR10 dataset}
\label{fig:resnet_exps}
\end{figure*}

\ifpaper
\fi

\ifpaper
\begin{figure*}[ht]
\centering     
\subfigure[CIFAR10]{\label{fig:val_logistic_cifar10_test_acc}\includegraphics[width=0.32\textwidth]{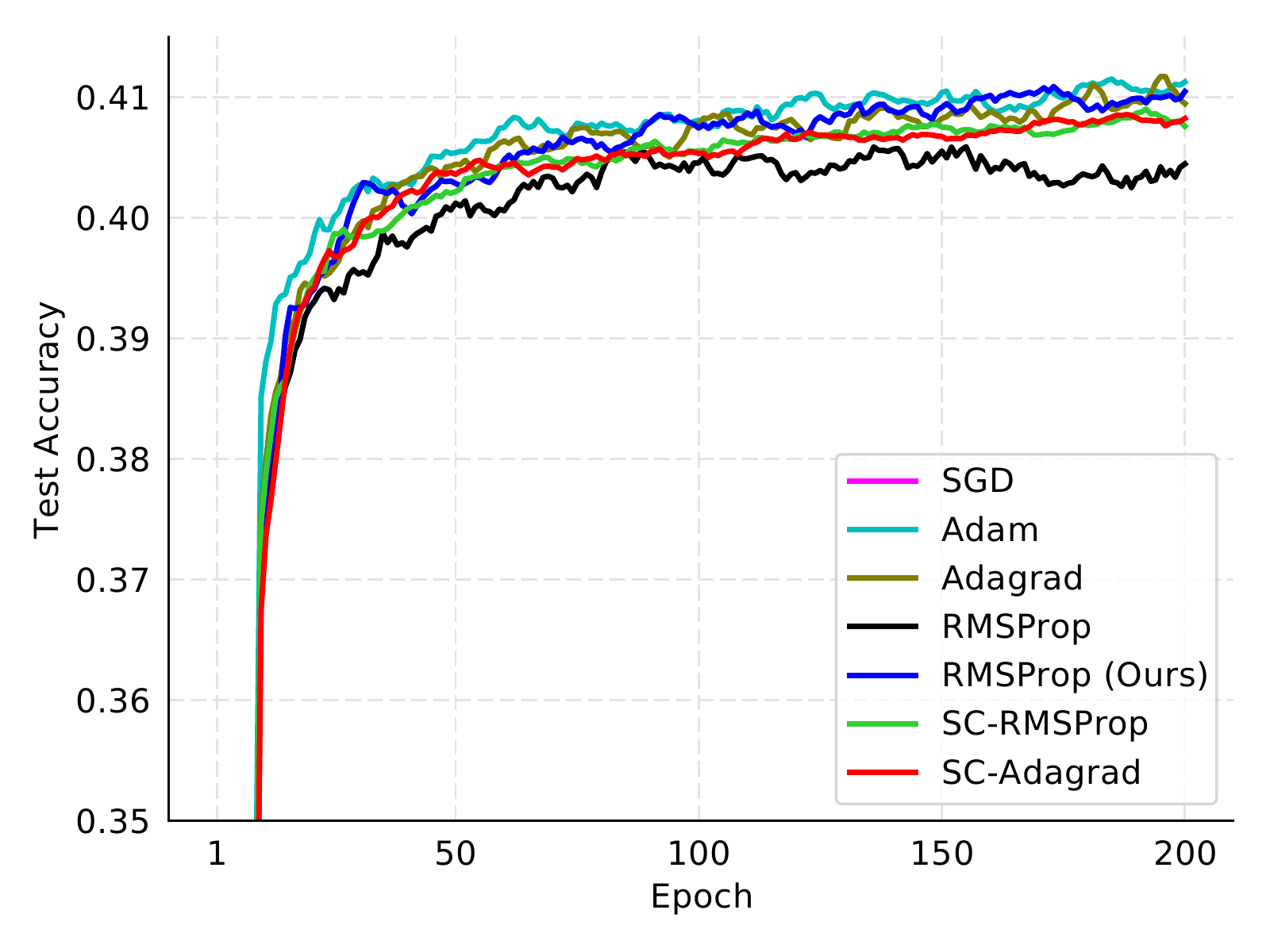}}
\subfigure[CIFAR100]{\label{fig:val_logistic_cifar100_test_acc}\includegraphics[width=0.32\textwidth]{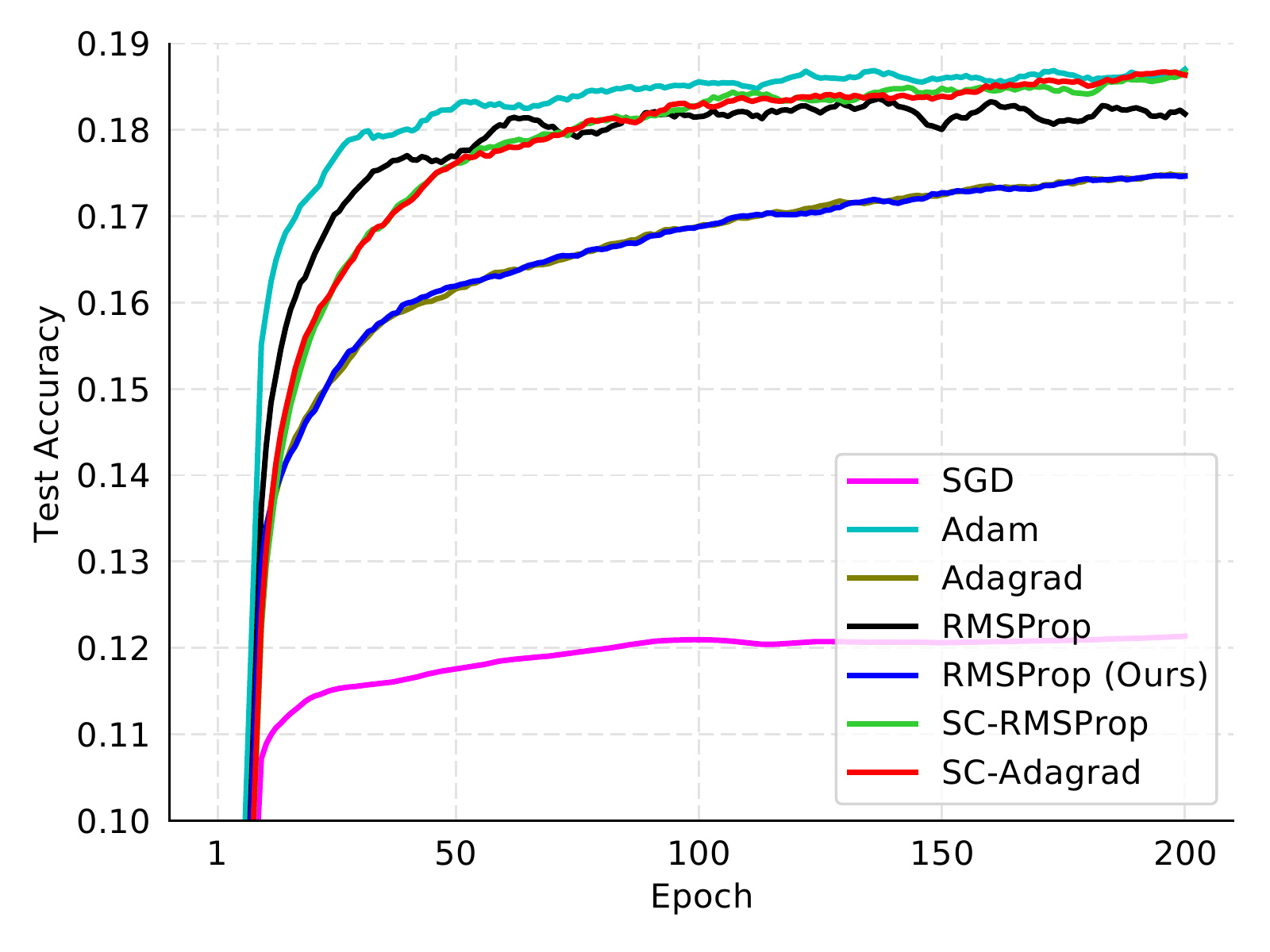}}
\subfigure[MNIST]{\label{fig:val_logistic_mnist_test_acc}\includegraphics[width=0.32\textwidth]{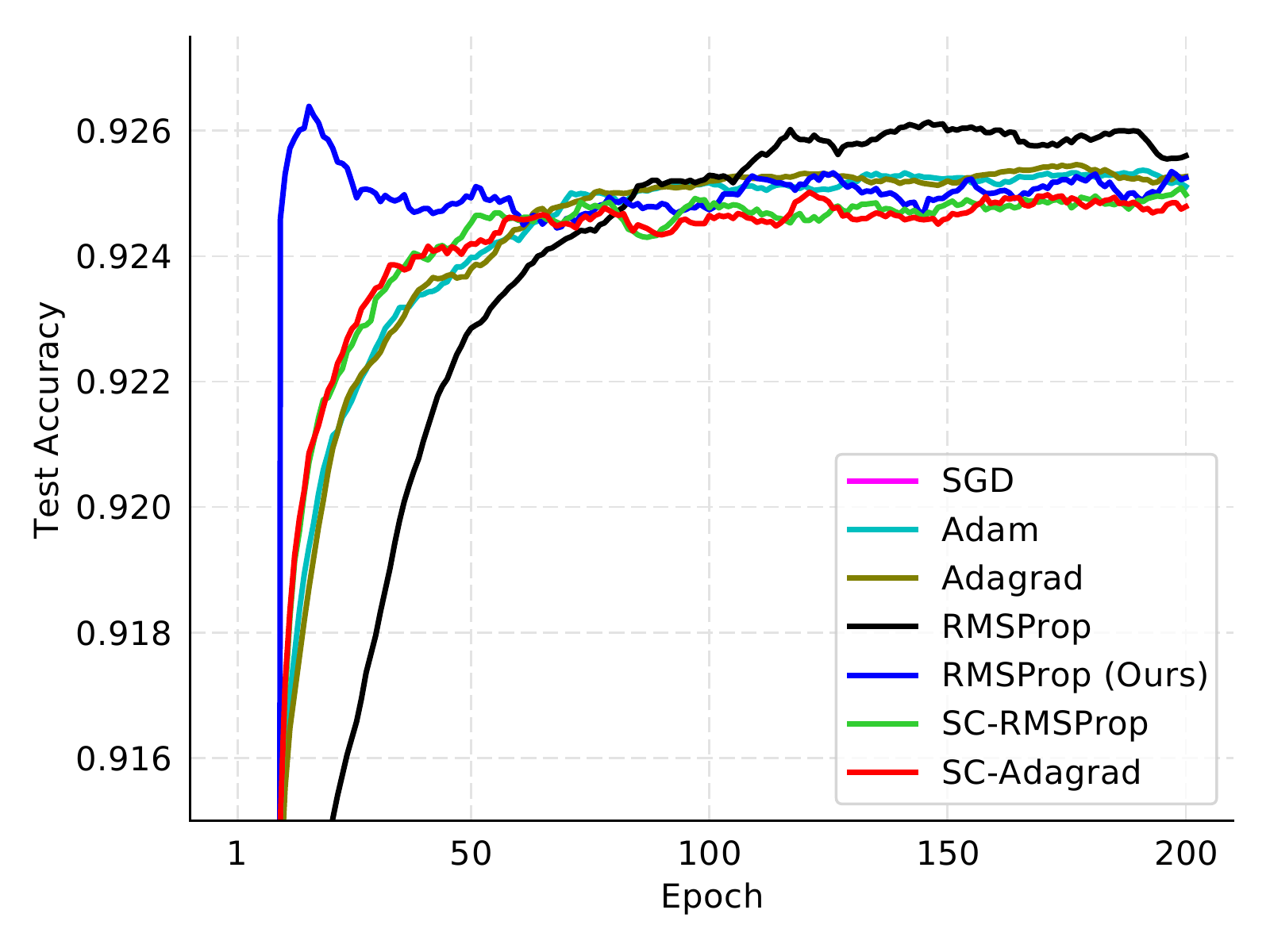}}
\caption{Test Accuracy vs Number of Epochs for L2-Regularized Softmax Regression}
\label{fig:val_logistic_test_acc}
\end{figure*}
\fi
\ifpaper
\begin{figure*}
\centering     
\subfigure[CIFAR10]{\label{fig:cnn_cifar10}\includegraphics[width=0.32\textwidth]{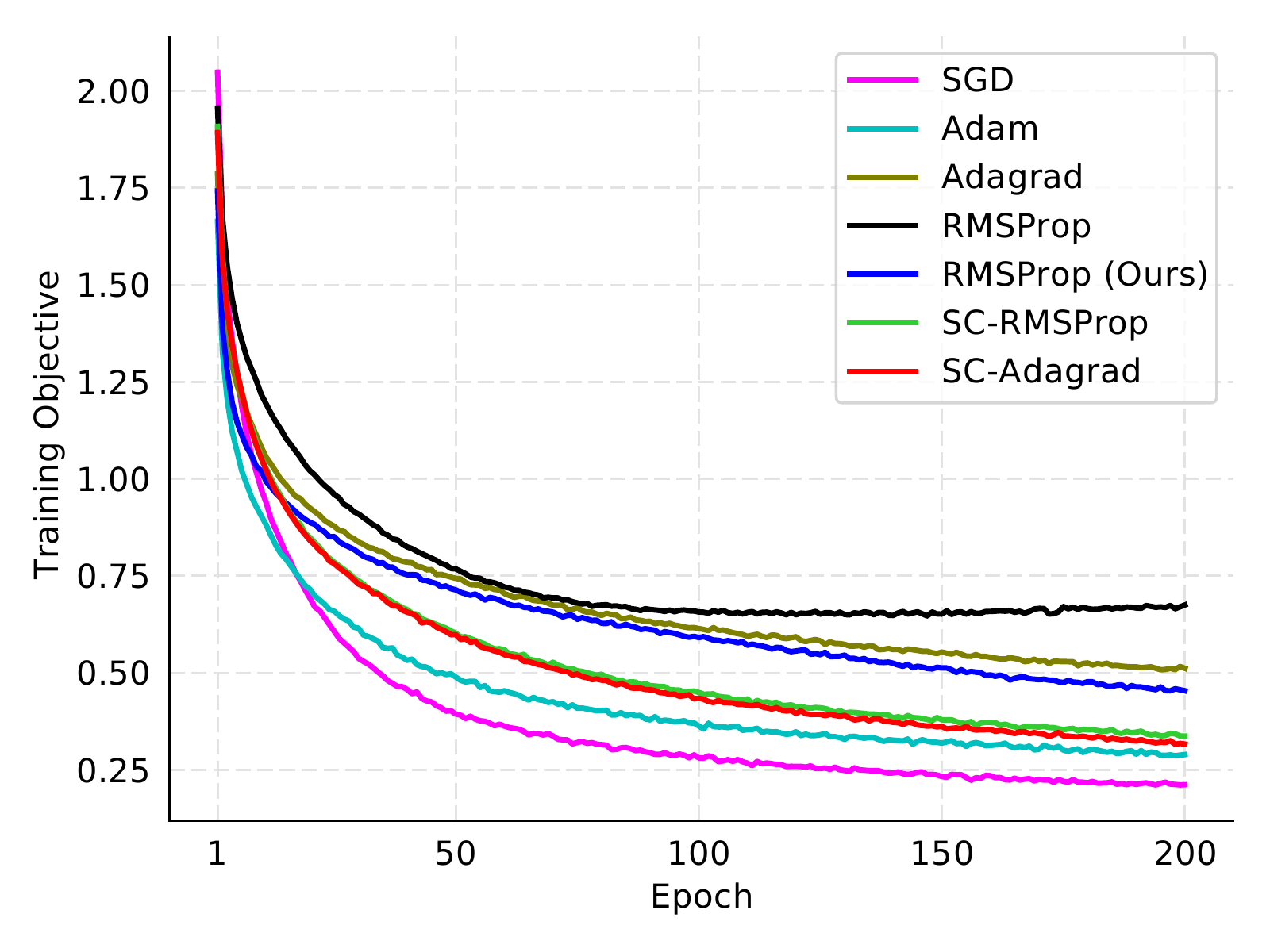}}
\subfigure[CIFAR100]{\label{fig:cnn_cifar100}\includegraphics[width=0.32\textwidth]{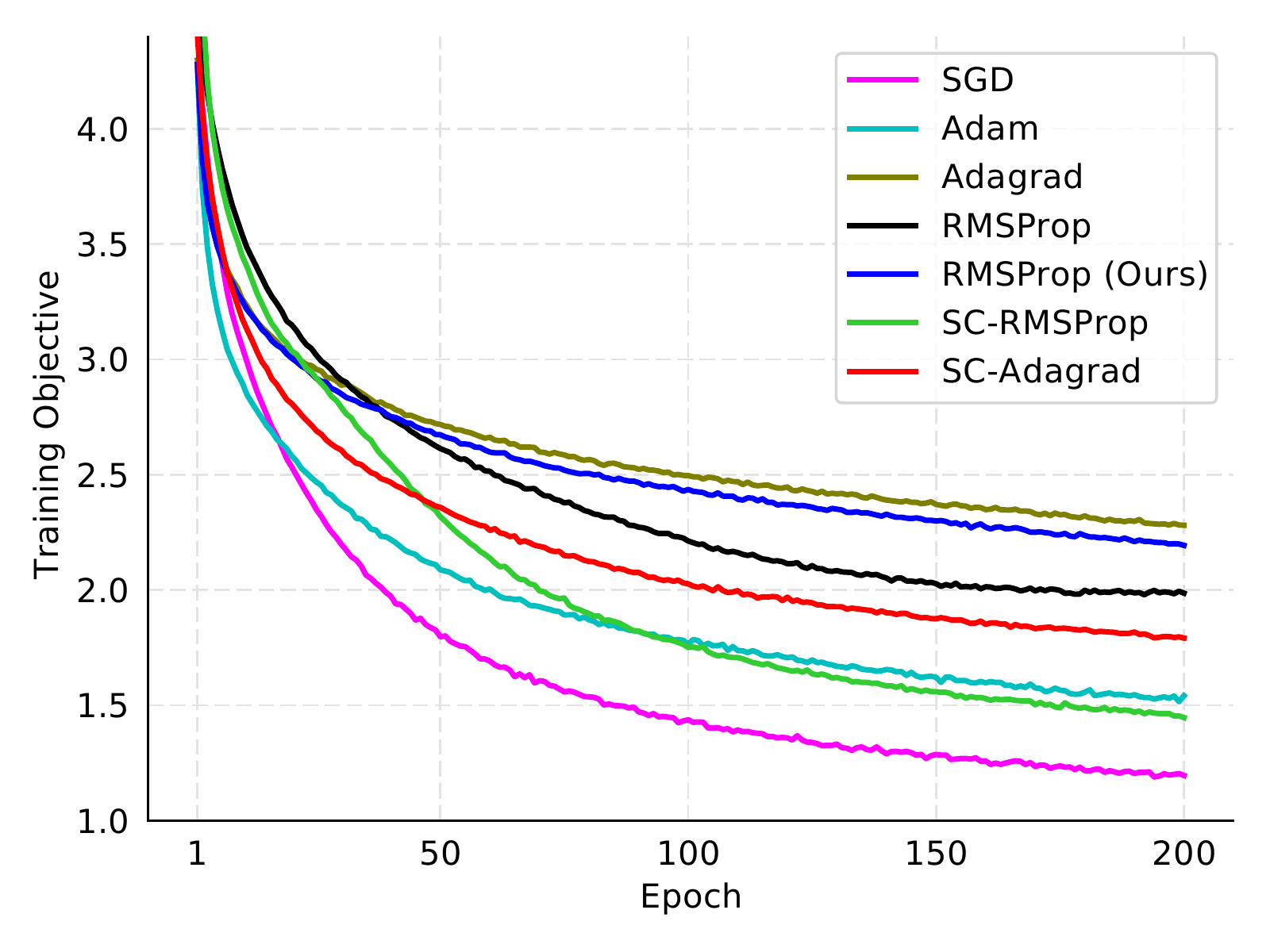}}
\subfigure[MNIST]{\label{fig:cnn_mnist}\includegraphics[width=0.32\textwidth]{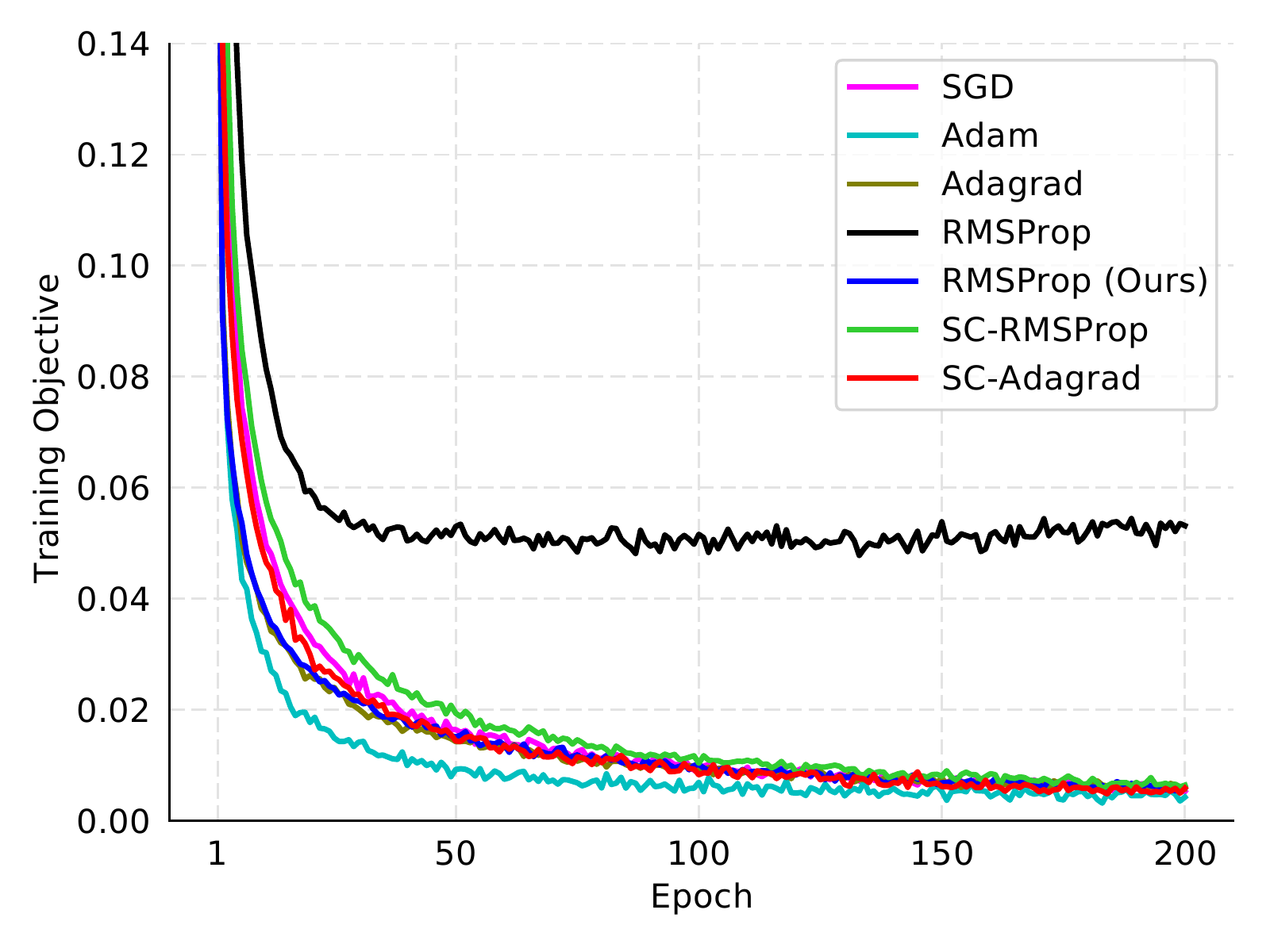}}
\caption{Training Objective vs Number of Epoch for 4-layer CNN}
\label{fig:cnn_test}
\end{figure*}
\fi
\ifpaper
\begin{figure*}
\centering     
\subfigure[CIFAR10]{\label{fig:mlp_cifar10}\includegraphics[width=0.32\textwidth]{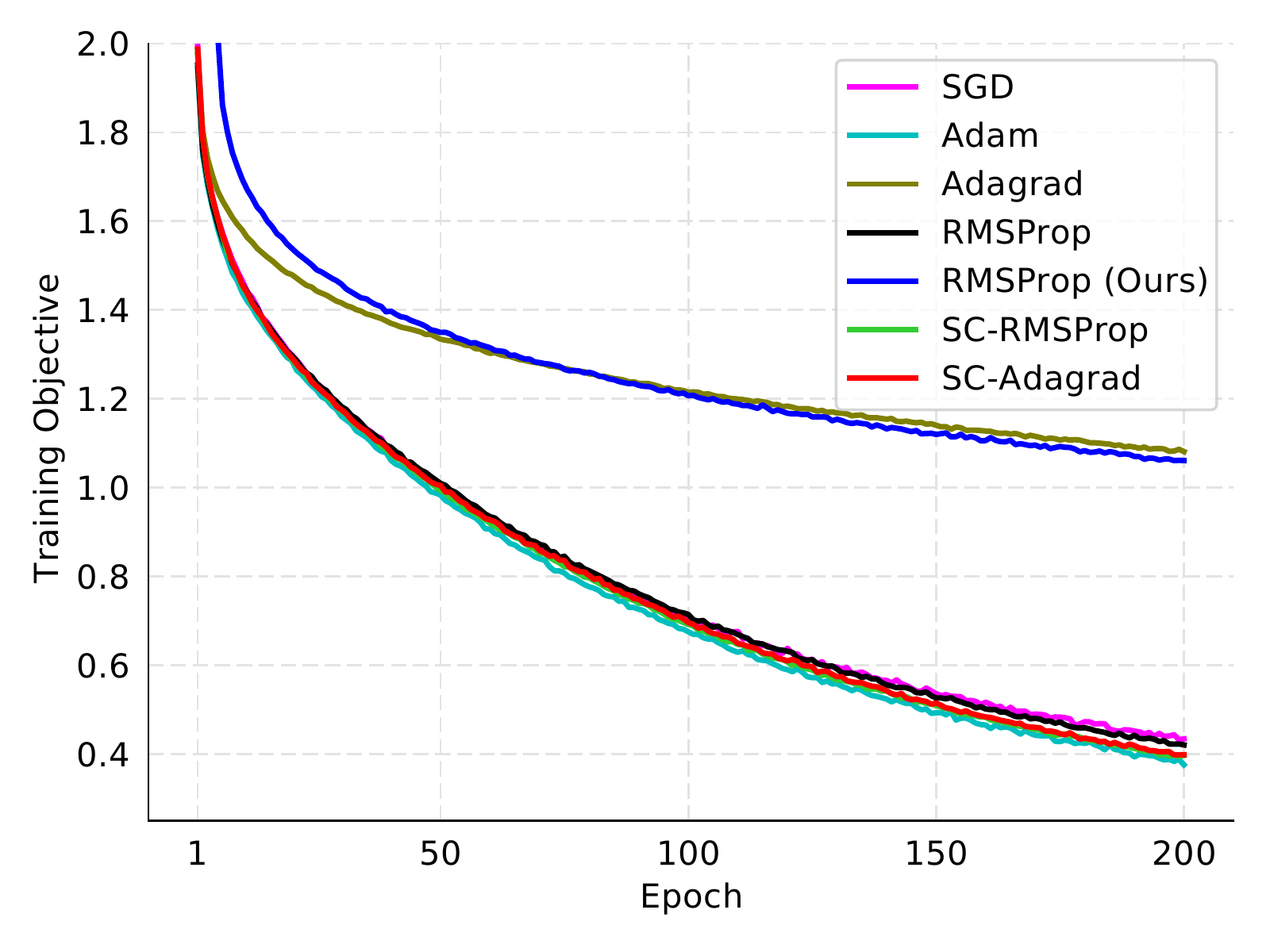}}
\subfigure[CIFAR100]{\label{fig:mlp_cifar100}\includegraphics[width=0.32\textwidth]{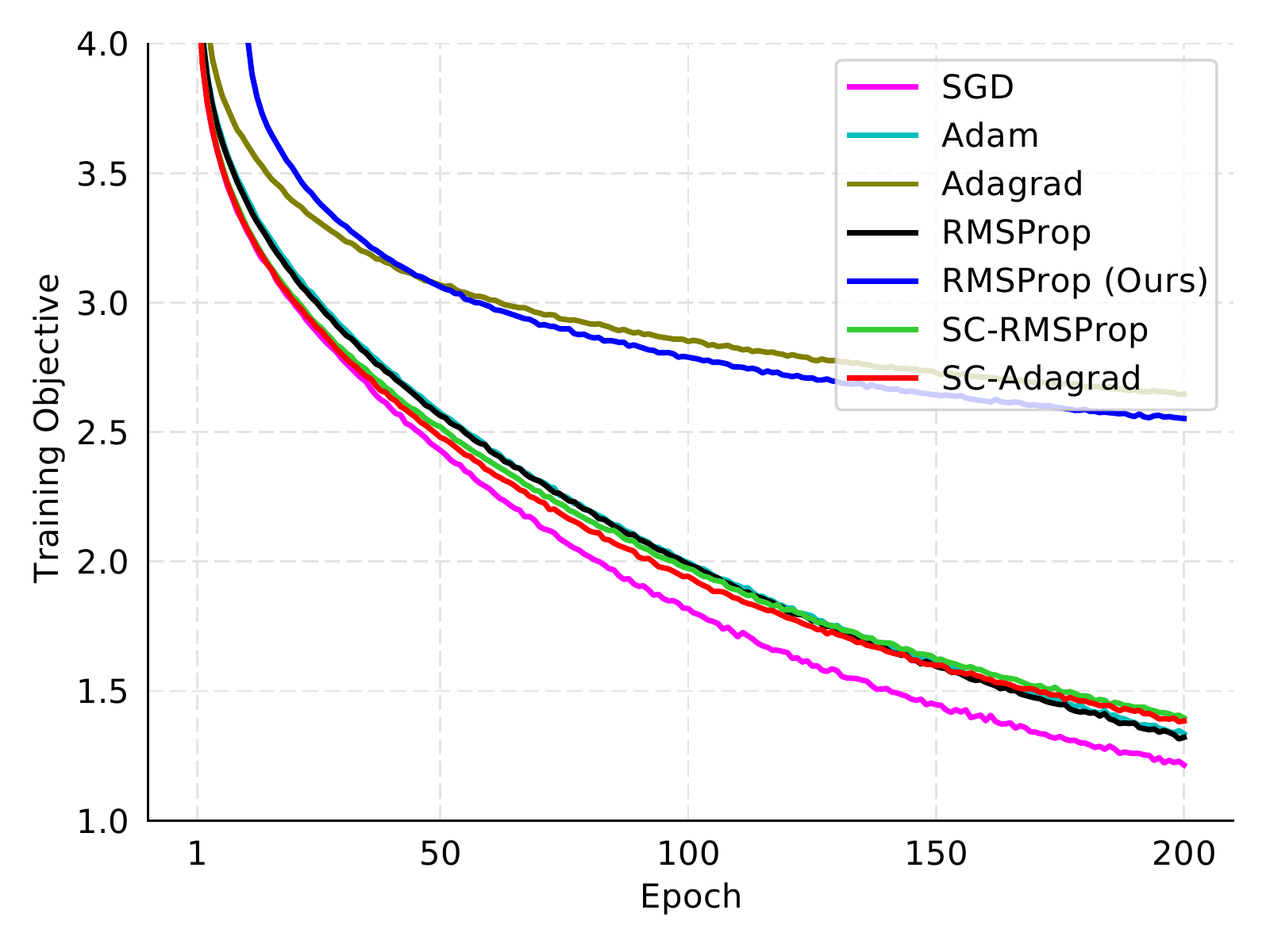}}
\subfigure[MNIST]{\label{fig:mlp_mnist}\includegraphics[width=0.32\textwidth]{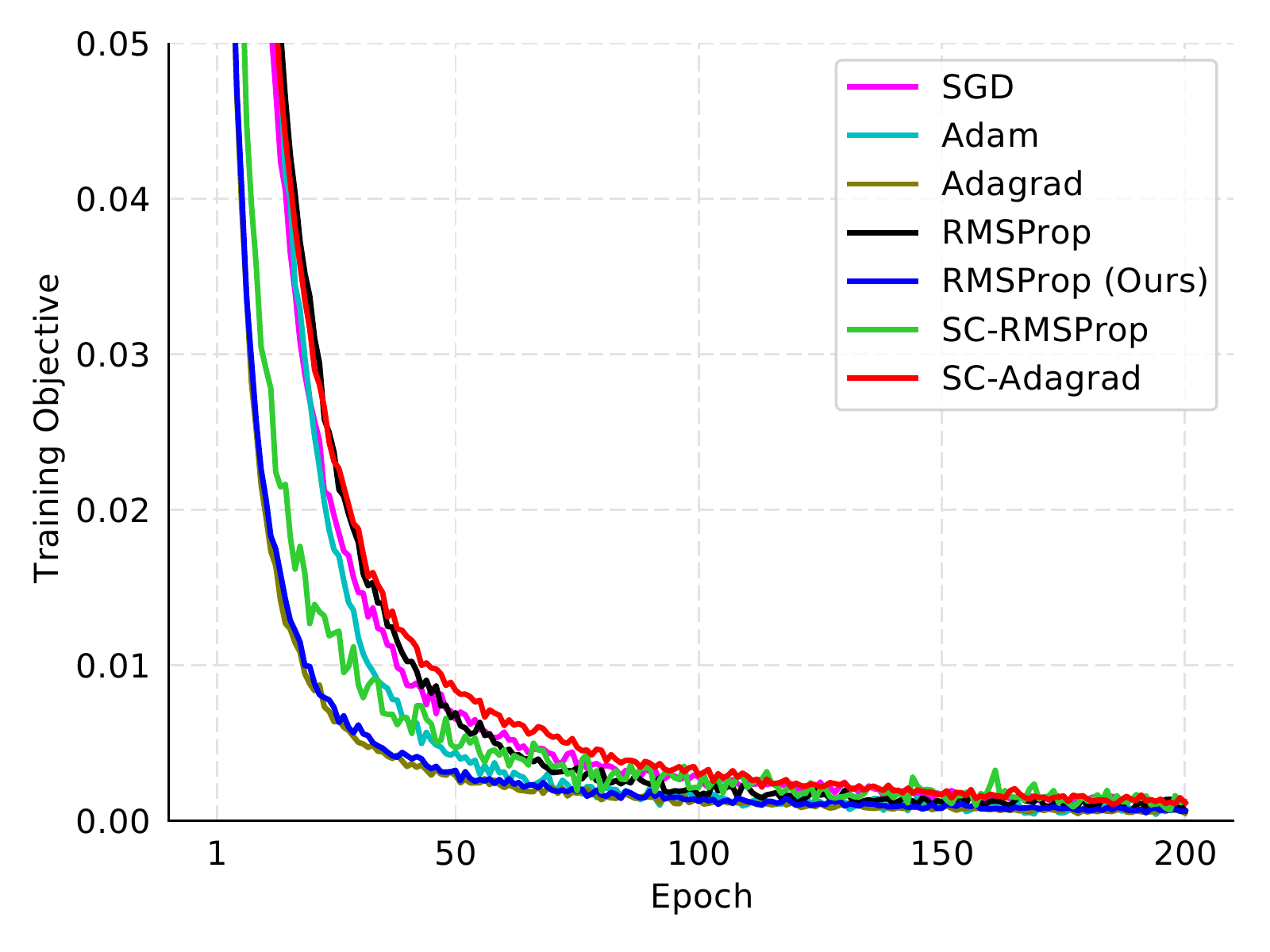}}
\vspace{-0.3cm}
\caption{Training Objective vs Number of Epoch for 3-layer MLP}
\label{fig:mlp_test}
\end{figure*}
\fi
\ifpaper
\begin{figure*}
\centering     
\subfigure[CIFAR10]{\label{fig:val_mlp_cifar10_test_acc}\includegraphics[width=0.32\textwidth]{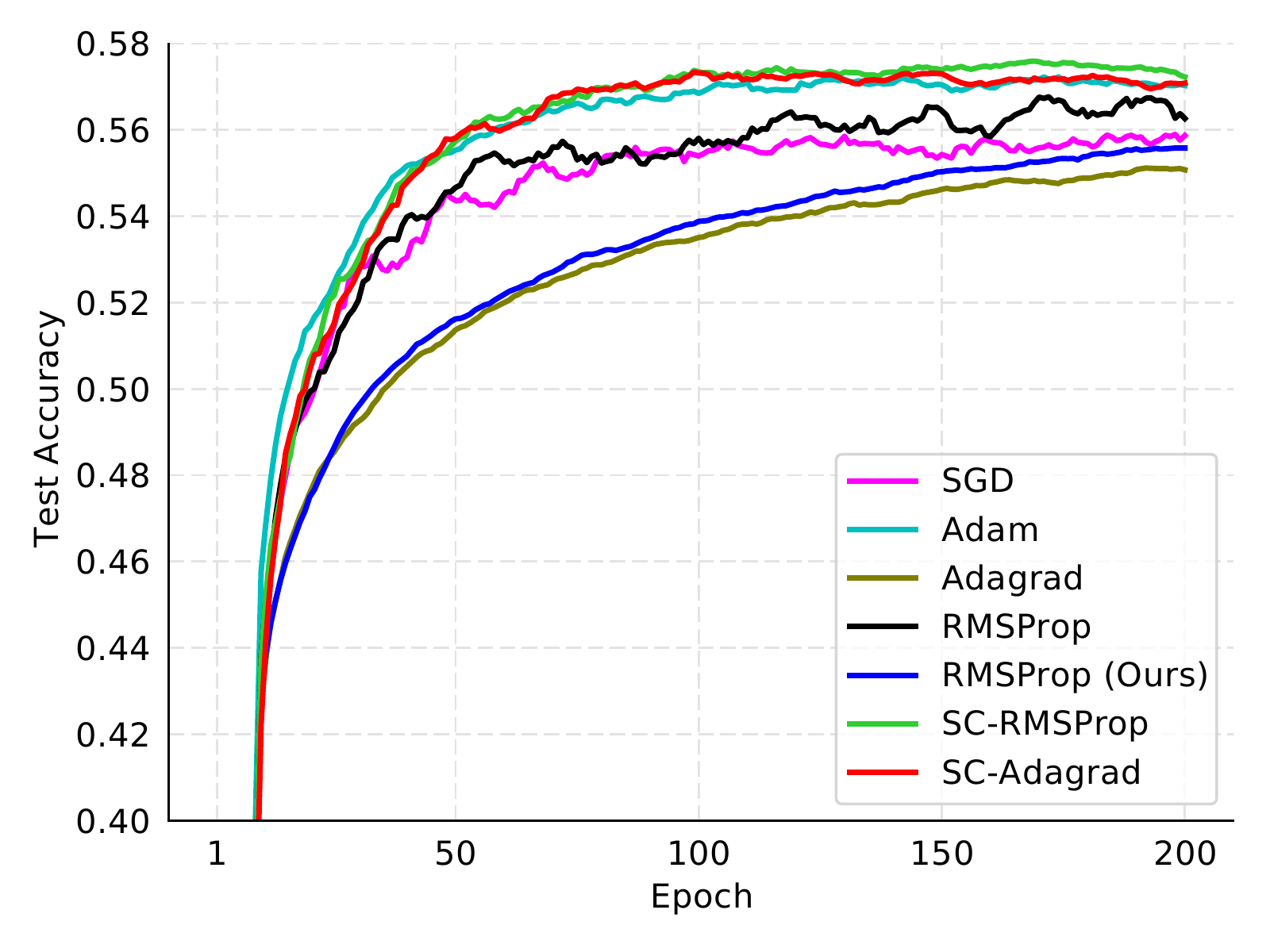}}
\subfigure[CIFAR100]{\label{fig:val_mlp_cifar100_test_acc}\includegraphics[width=0.32\textwidth]{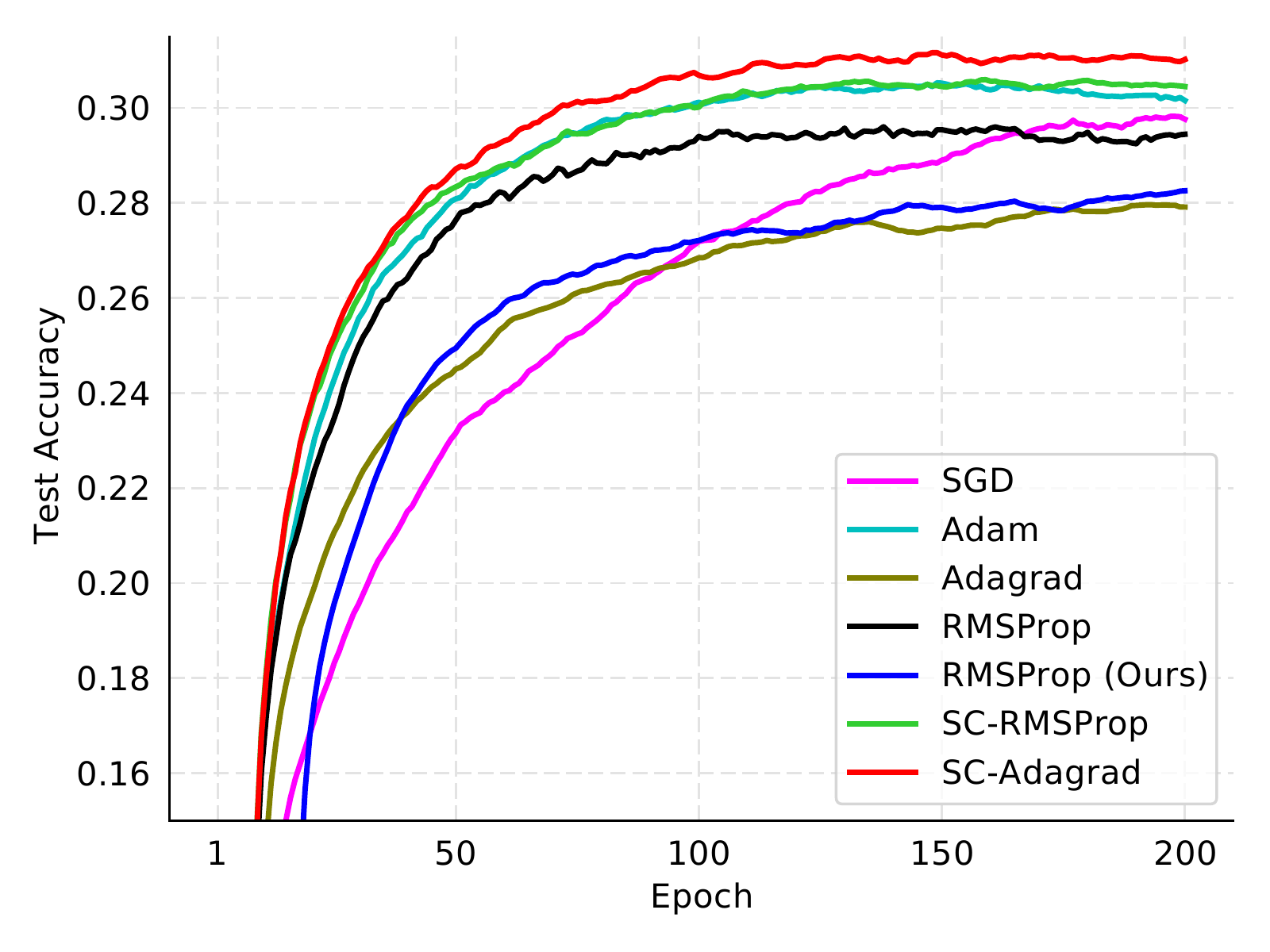}}
\subfigure[MNIST]{\label{fig:val_mlp_mnist_test_acc}\includegraphics[width=0.32\textwidth]{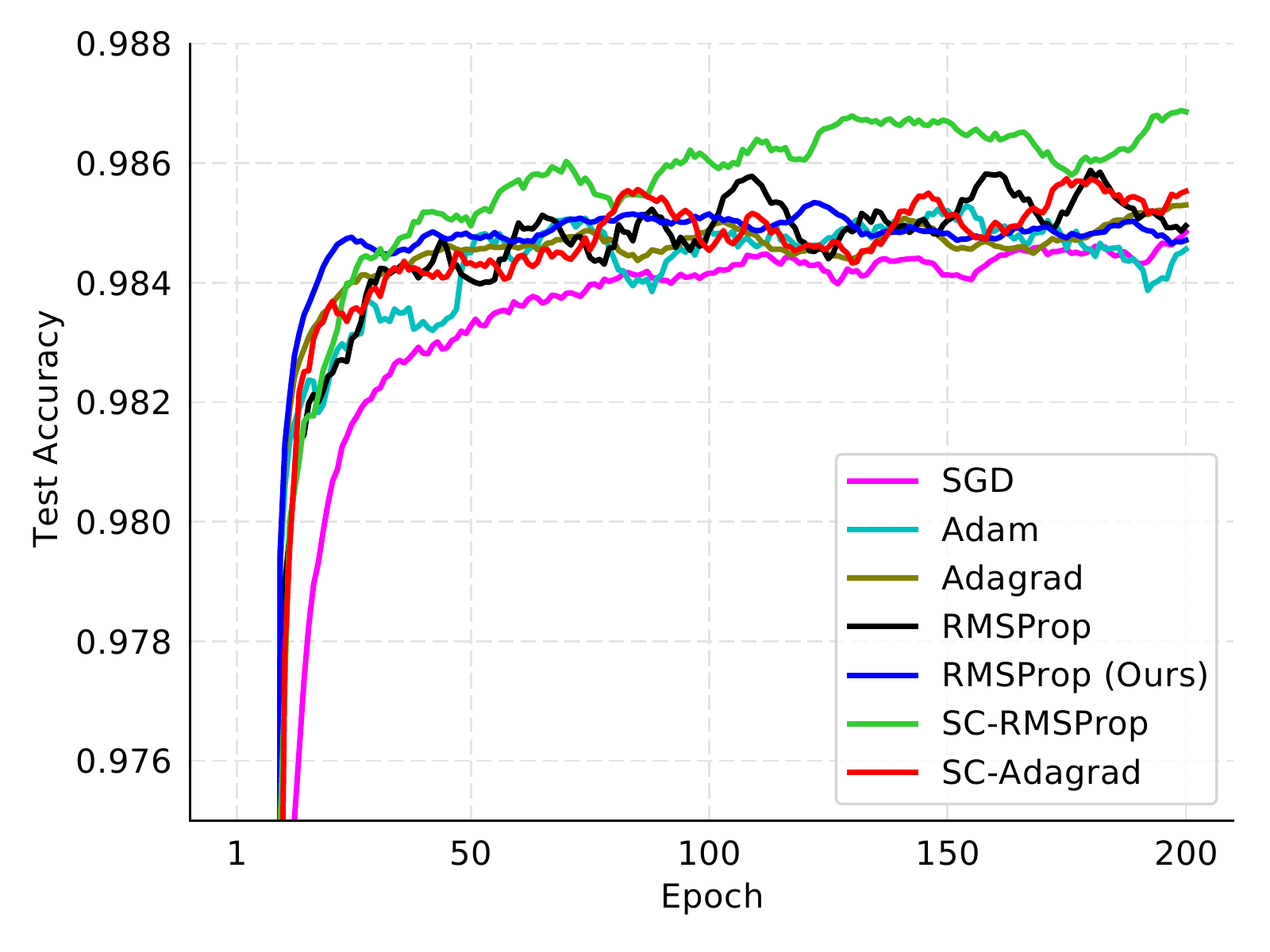}}
\vspace{-0.3cm}
\caption{Test Accuracy vs Number of Epochs for 3-layer MLP}
\label{fig:val_mlp_test_acc}
\end{figure*}
\fi
\section{Experiments}

The idea of the experiments is to show that the proposed algorithms are useful for standard learning problems in both online and batch settings. We are aware of the fact that in the strongly convex case online to batch conversion is not tight \cite{hazan2014beyond}, however that does not necessarily imply that the algorithms behave generally suboptimal. We compare all algorithms for a strongly convex problem and present relative suboptimality plots, $\log_{10}\Big(\frac{f(x_t)-p^*}{p^*}\Big)$, where $p^*$ is the global optimum, as well as separate regret plots, where we compare to the best optimal parameter in hindsight for the fraction of training points seen so far. 
On the other hand RMSProp was 
originally developed by  \citep{hinton2012lecture} for usage in deep learning. As discussed before the fixed choice of
$\beta_t$ is not allowed if one wants to get the optimal $O(\sqrt{T})$ regret bound in the convex case. Thus we think it is of interest to the deep learning community, if the insights from the convex optimization case transfer to deep learning. Moreover, 
Adagrad and RMSProp are heavily used in deep learning and thus it is interesting to compare their counterparts SC-Adagrad
and SC-RMSProp developed for the strongly convex case also in deep learning. 
\ifpaper
For the deep learning experiments we optimize the learning rate once for smallest training objective as well as for best test performance after a fixed number of epochs (typically 200 epochs).
\fi 

\textbf{Datasets:}
We use three datasets where it is easy, difficult and very difficult to achieve good test performance, just in order to see if this
influences the performance. For this purpose we use MNIST (60000 training samples, 10 classes), CIFAR10 (50000 training
samples, 10 classes) and CIFAR100 (50000 training samples, 100 classes). We refer to \cite{krizhevsky2009learning} for 
more details on the CIFAR datasets. 

\textbf{Algorithms:}
We compare 1) Stochastic Gradient Descent (SGD) \cite{bottou2010large} with $O(1/t)$ decaying step-size for the strongly convex problems and for non-convex problems we use a constant learning rate, 2) Adam \citep{kingma2014adam} , is used with step size decay of $\alpha_t = \frac{\alpha}{\sqrt{t}}$ for strongly convex problems and for non-convex problems we use a constant step-size.
3) Adagrad, see Algorithm \ref{alg:adagrad}, remains the same for strongly convex problems and non-convex problems. 4) RMSProp as proposed in \citep{hinton2012lecture} is used for both strongly convex problems and non-convex problems with $\beta_t$ = 0.9 $\forall t\geq 1$. 5) RMSProp (Ours) is used with step-size decay of $\alpha_t = \frac{\alpha}{\sqrt{t}}$  and $\beta_t = 1- \frac{\gamma}{t}$. In order that the parameter range is similar to the original RMSProp (\citep{hinton2012lecture}) we fix as $\gamma = 0.9$ for all experiment (note that for $\gamma=1$ RMSProp (Ours) is equivalent to Adagrad), 6) SC-RMSProp is used with stepsize $\alpha_t = \frac{\alpha}{t}$ and $\gamma = 0.9$  as RMSProp (Ours) 7) SC-Adagrad is used with a constant stepsize $\alpha$. The decaying damping factor for both SC-Adagrad and SC-RMSProp is used with $\xi_1=0.1,\, \xi_2 = 1$ for convex problems and we use $\xi_1 = 0.1,\, \xi_2 = 0.1$ for non-convex deep learning problems. Finally, the numerical stability parameter $\delta$ used in Adagrad, Adam, RMSProp is set to $10^{-8}$ as it is typically recommended for these algorithms.

\textbf{Setup:}
Note that all methods have only one varying parameter: the stepsize $\alpha$ which we choose from the set of  $\{1,0.1,0.01,0.001,0.0001\}$ for all experiments. By this setup no method has an advantage just because it has more hyperparameters over which it can optimize. The optimal rate is always chosen for each algorithm separately so that one achieves either best training objective or best test performance after a fixed number of epochs.

\textbf{Strongly Convex Case - Softmax Regression:} Given the training data $(x_i,y_i)_{i \in [m]}$ and let $y_i \in [K]$. we fit a linear model with cross entropy loss and use as regularization the squared Euclidean norm of the weight parameters.
The objective is then given as
\begin{align*}
& J(\theta) =-\frac{1}{m}\sum_{i=1}^m\text{log}\Bigg(\frac{e^{\theta_{y_i}^Tx_i + b_{y_i}}}{\sum_{j=1}^Ke^{\theta_j^Tx_i + b_j}}\Bigg)   + \lambda\sum_{k=1}^K \|\theta_k\|^2
\end{align*}
All methods are initialized with zero weights. The regularization parameter was chosen so that one achieves the best prediction
performance on the test set. The results are shown in Figure \ref{fig:logistic_test}. We also conduct experiments in an online setting, where we restrict the number of iterations to the number of training samples. Here for all the algorithms, we choose the stepsize resulting in best regret value at the end. We plot the Regret (\,in log scale\,) vs dataset proportion seen, and as expected SC-Adagrad and SC-RMSProp outperform all the other methods across all the considered datasets. Also, RMSProp (Ours) has a lower regret values than the original RMSProp as shown in Figure \ref{fig:test_online_logistic_regret}.

\textbf{Convolutional Neural Networks:} Here we test a 4-layer CNN with two convolutional (32 filters of size $3 \times 3$) and one fully connected layer (128 hidden units followed by 0.5 dropout). The activation function is ReLU and after the last convolutional layer we use max-pooling over a $2\times 2$ window and 0.25 dropout.  The final layer is a softmax layer and the final objective is cross-entropy loss. This is a pretty simple standard architecture and we use it for all datasets. The results are shown in 
\ifpaper
Figures \ref{fig:val_cnn_test_acc}, \ref{fig:cnn_test}.
\fi
\ifpaper
SC-RMSProp is competitive in terms of training objective on all datasets though SGD achieves the best performance. SC-Adagrad is not very competitive and the reason seems to be that the numerical stability parameter is too small.  RMSProp diverges on CIFAR10 dataset whereas RMSProp (Ours) converges on all datasets and has similar performance as Adagrad in terms of training objective.
\fi
 Both RMSProp (Ours) and SC-Adagrad perform better than all the other methods in terms of test accuracy for CIFAR10 dataset. On both CIFAR100 and MNIST datasets SC-RMSProp is very competitive.

\ifpaper
\textbf{Multi-Layer Perceptron:}
We also conduct experiments for a 3-layer Multi-Layer perceptron with 2 fully connected hidden layers and a softmax layer according to the number of classes in each dataset. For the first two hidden layers we have 512 units in each layer with ReLU activation function and 0.2 dropout. The final layer is a softmax layer.  We report the results in Figures \ref{fig:mlp_test}, \ref{fig:val_mlp_test_acc}. On all the datasets, SC-Adagrad and SC-RMSProp perform better in terms of Test accuracy and also have the best training objective performance on CIFAR10 dataset. On MNIST dataset, Adagrad and RMSProp(Ours)  achieves best training objective performance however SC-Adagrad and SC-RMSProp eventually performs as good as Adagrad. Here, the performance is not as competitive as Adagrad, because the numerical stability decay parameter of SC-Adagrad and SC-RMSProp are too prohibitive.
\fi

\textbf{Residual Network:}
We also conduct experiments for ResNet-18 network proposed in  \cite{he2016deep} where the residual blocks are used with modifications proposed in  \cite{he2016identity} on CIFAR10 dataset.  We report the results in Figures \ref{fig:resnet_exps}. SC-Adagrad, SC-RMSProp and RMSProp (Ours) have the best performance in terms of test Accuracy and RMSProp (Ours) has the best performance in terms of training objective along with Adagrad. 

\ifpaper
Given these experiments, we think that SC-Adagrad, SC-RMSProp and RMSProp (Ours)  are valuable new adaptive gradient techniques for deep learning.
\fi

\section{Conclusion}

We have analyzed RMSProp originally proposed in the deep learning community in the framework of online
convex optimization. We show that the conditions for convergence of RMSProp for the convex case are different
than what is used by \cite{hinton2012lecture} and that this leads to better performance in practice. We also
propose variants SC-Adagrad and SC-RMSProp which achieve logarithmic regret bounds for the strongly convex case.
Moreover, they perform very well for different network models and datasets and thus they are an interesting alternative to existing adaptive gradient schemes. In the future we want to explore why these algorithms perform so well in deep learning tasks even though they have been designed for the strongly convex case.

\section*{Acknowledgements}
We would like to thank Shweta Mahajan and all the reviewers for their insightful comments.

\bibliography{example_paper}
\bibliographystyle{icml2017}

\end{document}